\pdfoutput=1
\documentclass{article}

\usepackage[main, preprint]{neurips}

\usepackage[utf8]{inputenc} %
\usepackage[T1]{fontenc}    %
\usepackage{hyperref}       %
\usepackage{url}            %
\usepackage{booktabs}       %
\usepackage{amsfonts}       %
\usepackage{nicefrac}       %
\usepackage{microtype}      %
\usepackage{xcolor}         %

\usepackage{adjustbox}        %
\usepackage{algorithm}        %
\usepackage{algpseudocode}    %
\usepackage{amsfonts}         %
\usepackage{amsmath}          %
\usepackage{amssymb}          %
\usepackage{bm}               %
\usepackage{booktabs}         %
\usepackage{colortbl}         %
\usepackage{comment}          %
\usepackage{dsfont}           %
\usepackage{enumitem}         %
\usepackage{hyperref}         %
\usepackage{listings}         %
\usepackage{lscape}           %
\usepackage{makecell}         %
\usepackage{mathtools}        %
\usepackage{multirow}         %
\usepackage{multicol}         %
\usepackage{nicefrac}         %
\usepackage{pifont}           %
\usepackage{scalerel}         %
\usepackage{soul}             %
\usepackage{subcaption}       %
\usepackage{tabularx}         %
\usepackage{tablefootnote}    %
\usepackage{tcolorbox}        %
\usepackage{url}              %
\usepackage{wrapfig}          %
\usepackage{xcolor}           %
\usepackage{xspace}           %
\usepackage{xurl}             %
\usepackage{tabularray}
\usepackage{placeins}

\definecolor{llmoutputcolor}{rgb}{0.1,0.5,0.1}

\lstdefinestyle{llmstyle}{
  frame=single,
  basicstyle=\ttfamily\small\color{llmoutputcolor},
  breaklines=true,
  breakatwhitespace=false,
  columns=fullflexible
}

\usepackage{algorithm}
\usepackage{algpseudocode}

\algnewcommand\algorithmicparameters{\textbf{Hyperparameters:}}
\algnewcommand\Parameters{\item[\algorithmicparameters]}

\algnewcommand\algorithmicasync{\textbf{async}}
\algnewcommand\algorithmicawait{\textbf{await}}
\algnewcommand\Async{\algorithmicasync\ }
\algnewcommand\Await{\algorithmicawait\ }
\algnewcommand{\LineComment}[1]{\Statex \hspace{\algorithmicindent} $\triangleright$ #1}

\definecolor{myblue}{HTML}{8B98ED}
\newcommand{\fasymbol}{\textcolor{myblue}{*}}
\newcommand{\fa}{\textsuperscript{\fasymbol}}
\newcommand{\github}{\raisebox{-1.5pt}{\includegraphics[height=1.05em]{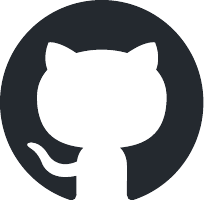}}\xspace}
\newcommand\blankfootnote[1]{%
  \let\thefootnote\relax\footnotetext{#1}%
  \let\thefootnote\svthefootnote%
}
\definecolor{tealbg}{RGB}{200,255,255}   %
\newcommand{\labelbox}[1]{%
  \colorbox{tealbg}{\textsc{\textcolor{teal}{#1}}}%
}

\newcommand{\ssc}[1]{{\small\textsc{#1}}}
\newcommand{\bsc}[1]{{\sc \textbf{#1}}\xspace}
\newcommand{\bssc}[1]{{\small \sc \textbf{#1}}\xspace}

\newcommand{\fssc}[1]{{\scriptsize \sc #1}\xspace}
\newcommand{\method}{\ssc{PRISM}\xspace}
\newcommand{\deepthink}{\ssc{DeepThink}\xspace}
\newcommand{\prm}{\ssc{PRM}\xspace}

\title{
\raisebox{-0.3em}{\includegraphics[height=1.8em]{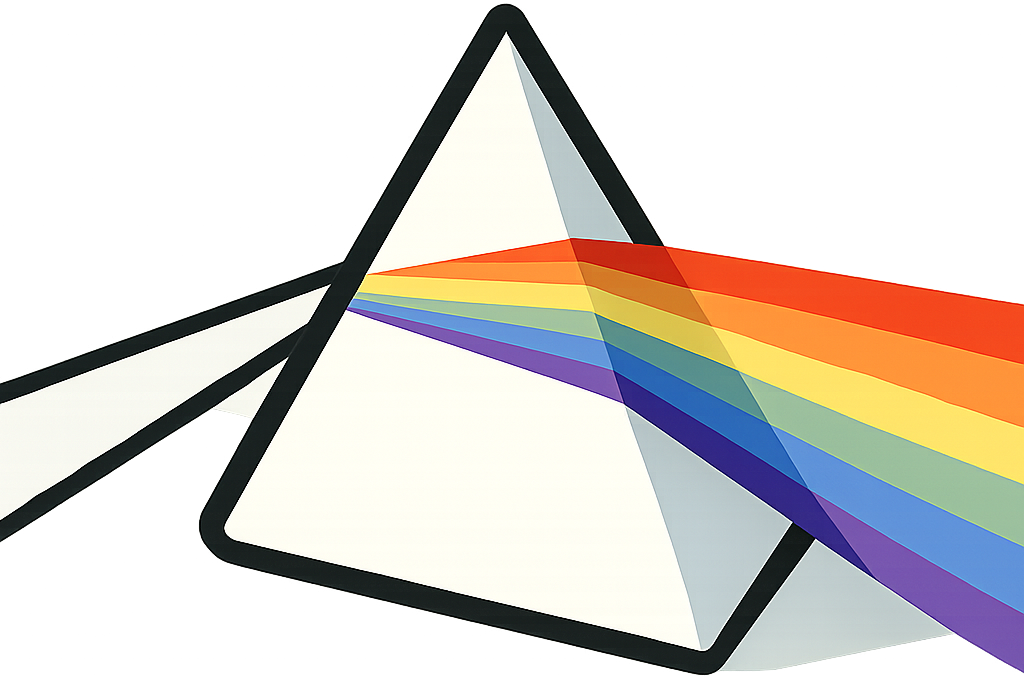}}%
  \hspace{0.3em}
PRISM: Pushing the Frontier of Deep Think via Process Reward Model-Guided  Inference}

\author{%
  Rituraj Sharma$^\fa$\textsuperscript{\includegraphics[height=1em]{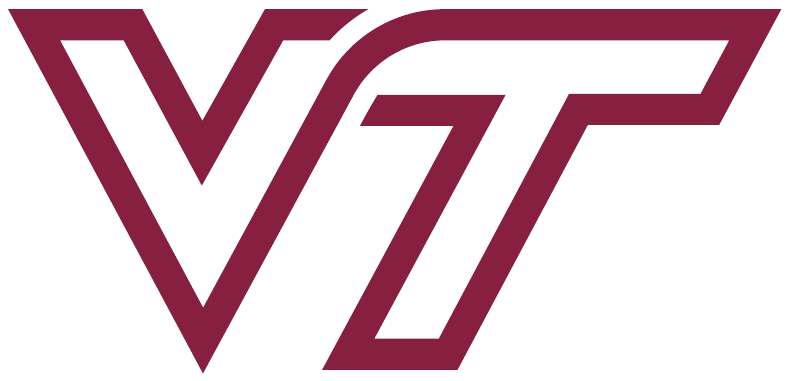}} \quad Weiyuan Chen$^\fa$\textsuperscript{\includegraphics[height=1em]{figures/vt.pdf}} \quad
  Noah Provenzano\textsuperscript{\includegraphics[height=1em]{figures/vt.pdf}} \quad
  Tu Vu\textsuperscript{\includegraphics[height=1em]{figures/vt.pdf}}\protect\\[0.75ex]
  \textsuperscript{\includegraphics[height=1em]{figures/vt.pdf}}Virginia Tech\protect\\[0.75ex]
\texttt{\{rituraj,cweiyuan9,noahpro,tuvu\}@vt.edu}\protect\\
\\
\github \url{https://github.com/Rituraj003/PRISM/}
\vspace{-12pt}
}

\begin{document}

\maketitle

\begin{abstract}
\deepthink methods improve reasoning by generating, refining, and aggregating populations of candidate solutions, which enables strong performance on complex mathematical and scientific tasks. However, existing frameworks can still suffer from unreliable correctness signals during inference, which creates a population-enhancement bottleneck where deeper deliberation amplifies errors, suppresses correct minority solutions, and yields weak returns to additional compute. In this paper, we introduce a functional decomposition of \deepthink systems and propose \method, a Process Reward Model (\prm)-guided inference algorithm that uses step-level verification to guide resampling, proposal acceptance, and solution aggregation. During refinement, \method uses \prm scores to guide population resampling and stochastic local refinement, which concentrates computation on higher-quality reasoning while preserving diversity. When paired with \prm-score-vote, a \prm-guided aggregation method, \method is competitive with or outperforms existing \deepthink methods, reaching 90.0\%, 75.4\%, and 71.4\% with \texttt{gpt-oss-20b} on AIME25, HMMT25, and GPQA Diamond, respectively, while exceeding \texttt{gpt-oss-120b}. Additionally, our analysis shows that \method produces consistent net-directional correction during refinement and maintains competitive accuracy when the initial population contains few correct candidates.

\blankfootnote{\textsuperscript{\fasymbol}Co-first authors.}
\end{abstract}

\begin{figure}[ht!]
\centering
\includegraphics[width=0.86\textwidth]{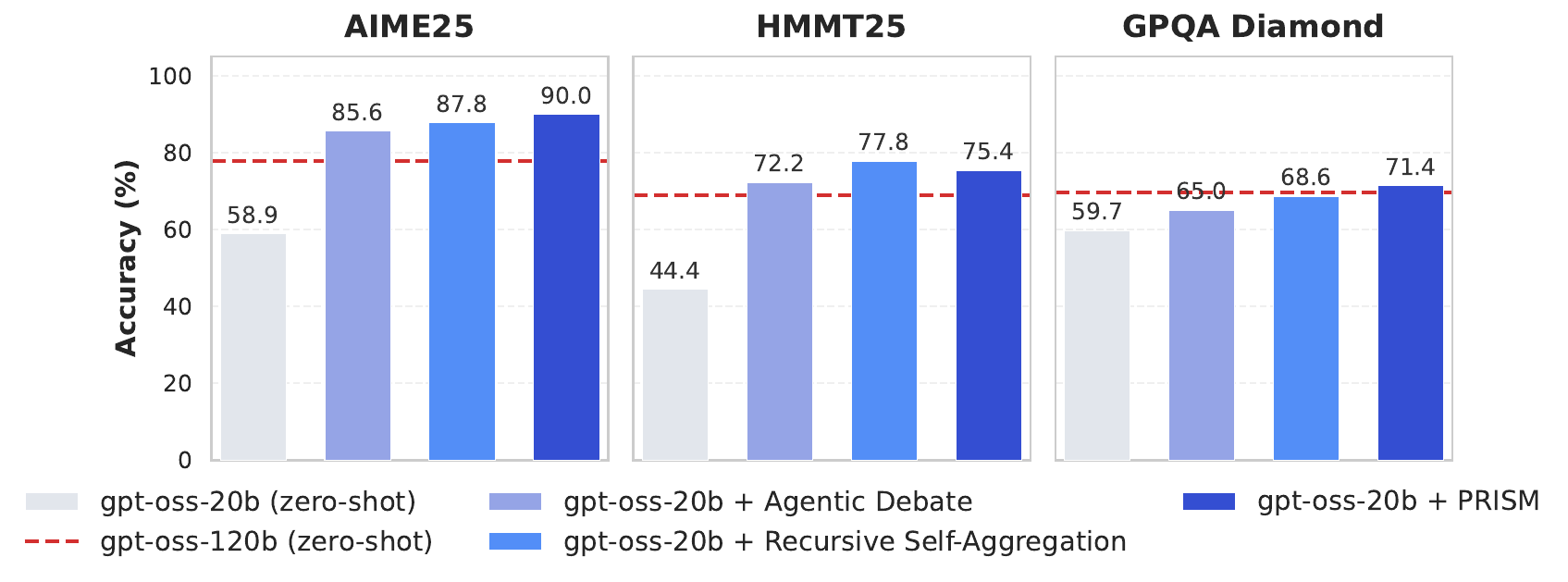}
\caption{\textbf{Accuracy on AIME25, HMMT25, and GPQA Diamond.} \method achieves competitive or superior performance relative to state-of-the-art \deepthink methods, enabling \texttt{gpt-oss-20b} to exceed \texttt{gpt-oss-120b} (see Table~\ref{tab:final_accuracy} for more results). See Section~\ref{sec:coding_generation} for coding benchmark results.}
\label{fig:figure1}
\end{figure}

\section{Introduction}
Recent advances in large language models (\ssc{LLMs}) have led to the emergence of \deepthink, a reasoning paradigm that allocates additional inference-time compute to simultaneously explore and combine multiple candidate solutions before producing a final answer~\citep{gemini25deepthink,gemini3deepthink}. \deepthink systems have demonstrated state-of-the-art performance on rigorous reasoning benchmarks and have reached gold-medal standards in competitions such as the International Mathematical Olympiad~\citep{luong2025advanced} and the International Collegiate Programming Contest~\citep{lin2025gemini}. Furthermore, these systems are increasingly used as research collaborators in mathematical and scientific discovery, where they support knowledge retrieval and rigorous verification~\citep{luong2026accelerating}.

Despite their successes, \deepthink systems remain poorly understood at a mechanistic level. Most existing frameworks are presented as monolithic pipelines, which makes it difficult to attribute improvements to specific design choices, diagnose failure modes, or understand the compute-performance tradeoff. Such a framing also obscures how components interact during extended reasoning, which limits systematic improvement and principled comparison across methods. This lack of transparency is concerning, since increased reasoning capability does not automatically translate into improved safety~\citep{caisaidashboard}.

\begin{figure*}[t!]
\centering
\includegraphics[width=0.85\textwidth]{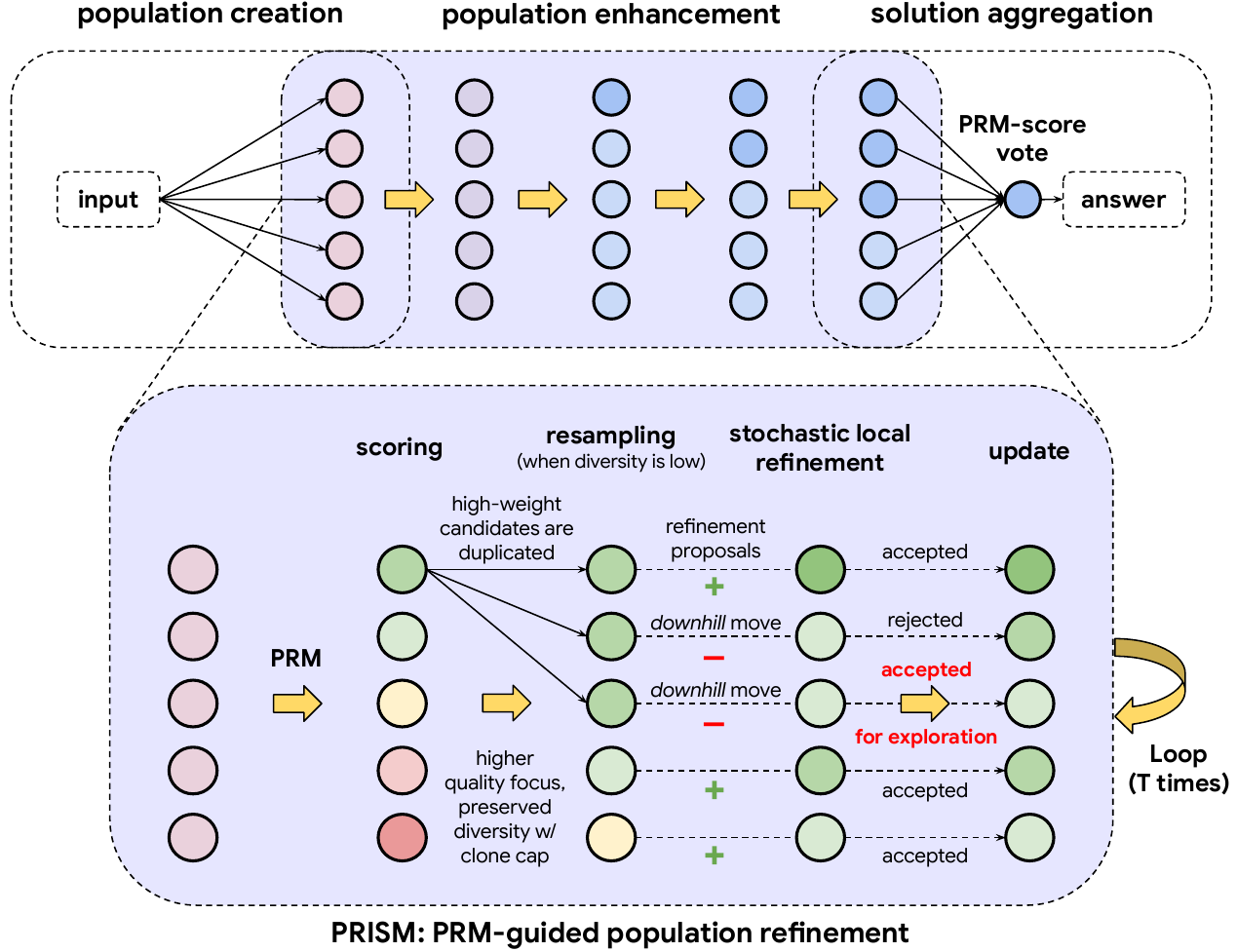}
\caption{\textbf{Functional taxonomy of \deepthink systems (\emph{top}) and overview of \method  (\emph{bottom}).} The top panel decomposes \deepthink into \emph{population creation}, \emph{population enhancement}, and \emph{solution aggregation}. The bottom panel illustrates \method's refinement mechanism, which uses Process Reward Model (\prm) scores to guide population resampling and stochastic local refinement.
}
\label{fig:prm_pipeline}
\vspace{-7pt}
\end{figure*}

To address this gap, we introduce a functional taxonomy of \deepthink frameworks that decomposes existing methods into three stages: \textbf{\textit{population creation}}, which generates diverse candidate solutions; \textbf{\textit{population enhancement}} (also referred to as \textbf{\textit{population refinement}}), which iteratively refines and improves these candidates; and \textbf{\textit{solution aggregation}}, which selects the final answer (see Figure~\ref{fig:prm_pipeline}, \emph{top}). This decomposition enables systematic analysis and controlled comparison across methods (see Section~\ref{sec:dissecting_deepthink}).

Through this lens, we identify population enhancement as the primary bottleneck for improving accuracy and efficiency. Empirically, we observe that parallel sampling followed by majority voting often performs competitively with sophisticated \deepthink systems, which suggests that much of the performance arises from initial population diversity and aggregation rather than iterative refinement.

Our taxonomy further reveals key limitations of current refinement strategies. Many methods repeatedly rewrite entire solutions without a stable quality signal. As a result, increasing refinement depth (the number of refinement iterations) does not reliably improve population quality and can instead propagate or amplify errors. Majority-based approaches, which steer candidates toward the most frequent solution, provide partial guidance through consensus but introduce a complementary failure mode, in which infrequent yet correct trajectories are suppressed by more frequent but incorrect ones, a phenomenon we refer to as \emph{majority dilution}. Consequently, current systems often fail to fully exploit their generated populations and exhibit suboptimal compute-performance tradeoffs.

Motivated by these findings, we propose \method
(\textbf{P}rocess reward model-guided \textbf{R}efinement, \textbf{I}teration, and \textbf{S}election \textbf{M}echanisms),
an inference algorithm that %
introduces explicit step-level correctness signals into both population refinement and solution aggregation (Figure~\ref{fig:prm_pipeline}, \emph{bottom}). \method uses a Process Reward Model (\prm) to evaluate reasoning trajectories based on their internal steps, which provides quality feedback independent of how frequently a trajectory appears in the population. During refinement, candidate solutions are treated as particles in a \prm-defined energy landscape and iteratively updated via score-guided resampling and stochastic proposals, which concentrate probability mass on higher-quality reasoning while preserving exploration. Proposed refinements with higher \prm scores are more likely to be accepted, while lower-scoring proposals can still be accepted to preserve exploration. Taken together, these mechanisms turn iterative refinement into directional error correction.
The final prediction selects the candidate with the highest aggregate \prm score (see Section~\ref{sec:prism}).

Across rigorous reasoning benchmarks, \method is competitive with or outperforms state-of-the-art \deepthink methods, achieving 90.0\%, 75.4\%, and 71.4\% with \texttt{gpt-oss-20b} on AIME25, HMMT25, and GPQA Diamond, respectively, while exceeding \texttt{gpt-oss-120b} \citep{agarwal2025gpt}.
Beyond accuracy, \method mitigates key failure modes: it produces consistent directional error correction during refinement (positive \emph{NetFlip}, where incorrect solutions are corrected more often than correct ones are degraded) and remains effective in low-correctness regimes by preserving correct minority trajectories (see Section~\ref{sec:exp_ablations}).

\vspace{-2pt}
To summarize, our main contributions are:
\begin{itemize}
    \item \textbf{A functional taxonomy of \bssc{DeepThink} systems:} We decompose existing frameworks into population creation, enhancement, and aggregation, which sheds light on their internal mechanisms and identifies population enhancement as a primary bottleneck for improving reasoning performance.
    \item \textbf{New \bssc{DeepThink} method:} We propose \method, a \prm-guided inference algorithm that uses step-level correctness signals to transform iterative refinement into directional error correction and inform final solution aggregation.
    \item \textbf{Improved performance, robustness, and refinement behavior:} Across rigorous benchmarks, \method is competitive with or outperforms state-of-the-art \deepthink methods, produces consistent directional error correction during refinement, and remains reliable in low-correctness regimes.
\end{itemize}
\vspace{-2pt}
Taken as a whole, we hope that our work will spur more principled research into scalable and reliable inference-time reasoning.

\vspace{-5pt}
\section{Dissecting \bsc{DeepThink} frameworks}
\label{sec:dissecting_deepthink}

In this section, we introduce a functional decomposition of \deepthink  frameworks into population creation, enhancement, and aggregation. Through this lens, we identify a population-refinement bottleneck and present empirical observations that motivate our method introduced in Section~\ref{sec:prism}.

\vspace{-5pt}
\subsection{A functional taxonomy of \bsc{DeepThink} frameworks}
We decompose \deepthink frameworks into three stages: \textbf{\textit{population creation}} (generating candidate solutions), \textbf{\textit{population enhancement}} (iteratively refining candidates), and \textbf{\textit{solution aggregation}} (synthesizing the answer). Each framework can be understood as instantiating these stages through distinct design choices. This decomposition isolates the role of each stage in shaping reasoning quality and enables a mechanistic analysis of how additional compute translates into correctness.

\vspace{-5pt}
\paragraph{Population creation:} This stage generates an initial, diverse set of candidate reasoning trajectories. The most common approach produces multiple candidates independently via stochastic decoding (e.g., temperature or nucleus sampling~\citep{Holtzman2020The}), as in~\citet{wang2023selfconsistency}; we refer to this strategy as \textbf{Sample-N}. An alternative is single-pass multi-candidate generation, where the model produces several solutions within one generation, sometimes with associated plausibility signals (e.g., verbalized sampling~\citep{zhang2025verbalized}). Population creation primarily contributes diversity rather than correctness: it increases the probability that at least one correct trajectory exists but does not concentrate probability mass on correct solutions. Consequently, downstream performance depends critically on how effectively later stages identify and amplify high-quality candidates.

\vspace{-5pt}
\paragraph{Population enhancement:} This stage operates on the initially generated candidate population and iteratively refines candidates with the goal of increasing the density of correct reasoning trajectories while preserving useful diversity~\citep{choi2025debate}. Existing methods fall into several broad categories, including \emph{critic-guided refinement}, which uses a critic to verify and revise each candidate solution independently~\citep{chai2025scimaster}; \emph{multi-agent interaction}, where agents iteratively challenge and update one another's reasoning~\citep{wang2025retrievalaugmented}; \emph{consensus-driven updating}, which steers candidates toward the majority belief in the population~\citep{choi2025debate}; and \emph{aggregation-based refinement}, which recursively recombines candidate solutions to produce improved ones~\citep{venkatraman2025recursive}. Despite differing mechanisms, these methods share a common objective: to improve population quality across refinement depth.

\vspace{-2pt}
In practice, however, population enhancement is often the least reliable stage. Refinement can propagate early errors, suppress correct minority trajectories, or drift toward frequent but incorrect reasoning patterns. As a result, population quality often oscillates or degrades rather than improving monotonically with additional refinement, which creates a population-enhancement bottleneck. %

\vspace{-5pt}
\paragraph{Solution aggregation:} This stage converts the final candidate population into a single prediction. The most common mechanism selects the most frequent answer among candidates (Majority Vote). %
More expressive strategies use model-based aggregation, where an \ssc{LLM} synthesizes a final answer by conditioning on the entire population (\ssc{LLM} Aggregate). However, aggregation implicitly assumes that population enhancement has produced a sufficiently high-quality candidate set. When this assumption fails, aggregation can amplify noise rather than correct it, which further highlights the role of population enhancement in determining overall performance. In this work, we additionally explore \prm-guided correctness-aware aggregation (\prm-score Vote), which selects the candidate with the highest aggregate \prm score.
\vspace{-5pt}
\subsection{Empirical motivation: The population-refinement bottleneck}
\label{sec:empirical_motivation}
\deepthink frameworks implicitly rely on a \emph{monotonic refinement assumption}: additional refinement iterations should progressively improve population quality and translate into higher final accuracy. Under controlled conditions where population creation is \emph{fixed} and only enhancement and aggregation vary, we find that this assumption frequently breaks down.

\vspace{-2pt}
First, \emph{diversity combined with simple aggregation is already a strong baseline.} Parallel sampling followed by majority voting often achieves competitive performance at low compute, while several refinement-heavy frameworks yield only marginal gains despite substantially higher token usage. Second, \emph{increasing refinement depth does not reliably improve population quality.} Across settings, the fraction of correct candidates often oscillates and can degrade over successive refinement iterations, which indicates that refinement updates often behave as stochastic perturbations rather than directional improvements. Finally, \emph{aggregation performance depends critically on the output of population enhancement}. When the candidate population is noisy, aggregation mechanisms can reinforce or rationalize incorrect majorities instead of filtering them (see Section~\ref{sec:exp_ablations} for details).

\vspace{-2pt}
Taken as a whole, these findings reveal \emph{a population-refinement bottleneck}. Without a correctness-sensitive signal, refinement updates can preserve, propagate, or amplify errors, which limits the returns from additional inference-time compute and prevents systems from fully exploiting their candidate populations. Overcoming this bottleneck requires inference mechanisms that incorporate explicit correctness signals to guide both population refinement and aggregation.%

\vspace{-2pt}
\section{\bsc{PRISM: PRM}-guided Refinement, Iteration, and Selection Mechanisms}
\label{sec:prism}
We propose \method, a Process Reward Model (\prm)-guided inference algorithm that addresses the population-refinement bottleneck by injecting explicit step-level correctness signals into both refinement dynamics and final selection.

During refinement, \method maintains a population of candidate reasoning traces and uses \prm scores to guide their updates. Each refinement iteration performs three operations: \textbf{\textit{scoring}}, \textbf{\textit{resampling}}, and \textbf{\textit{stochastic refinement}}. It scores candidates using \prm step-level feedback, resamples when probability mass collapses onto a small subset of candidates, and accepts stochastic refinements probabilistically based on \prm scores. Algorithm~\ref{alg:prm-mcmc} summarizes the procedure.
\begin{algorithm}[t]
\caption{\textbf{PRM-guided Resampling and Informed Stochastic Mutation.}}
\label{alg:prm-mcmc}
\begin{algorithmic}[1]
\Require Initial population $\mathcal{P}^{(0)}=\{\tau_i^{(0)}\}_{i=1}^N$, verifier $V$ (\prm), iterator $I$, comparator $C$
\Parameters depth $T$, $T_{\text{smc}}$, ESS threshold $\alpha$, noise $\eta$, clamp $c$, clone cap $\kappa$
\Ensure Updated population $\mathcal{P}^{(T)}$
\For{$t=1$ \textbf{to} $T$}
  \Statex \labelbox{Scoring}
  \ForAll{$\tau_i\in\mathcal{P}^{(t-1)}$ \textbf{in parallel}}
    \State $\tau_i \leftarrow \ssc{StepwiseNormalize}(\tau_i)$
    \State $\text{fb}_i \leftarrow V(\tau_i)$; \ $s_i \leftarrow \textsc{Score}(\text{fb}_i)\in[0,1]$
    \State $w_i \leftarrow s_i^{1/T_{\text{smc}}}$
  \EndFor
  \State $(\mathcal{P}^{(t-1)},\{s_i\}) \leftarrow \ssc{Arbitrate}(\mathcal{P}^{(t-1)},\{s_i,\text{fb}_i\},C,c)$
  \State $w_i \leftarrow s_i^{1/T_{\text{smc}}}$

  \Statex \labelbox{Resampling}
  \State $\mathrm{ESS} \leftarrow \frac{(\sum_i w_i)^2}{\sum_i w_i^2}$
  \If{$\mathrm{ESS}<\alpha N$}
    \State $\mathcal{P}^{(t-1)} \leftarrow \ssc{SystematicResample}(\mathcal{P}^{(t-1)},\{w_i\})$
    \State $\mathcal{P}^{(t-1)} \leftarrow \ssc{CapCopies}(\mathcal{P}^{(t-1)},\lceil \kappa N\rceil)$
  \EndIf

  \Statex \labelbox{Stochastic local refinement}
  \ForAll{$\tau_i\in\mathcal{P}^{(t-1)}$ \textbf{in parallel}}
    \State $\tau'_i \leftarrow I(\tau_i,\text{fb}_i;\ \eta)$ \Comment{inject ``different approach'' w.p. $\eta$}
    \State $\text{fb}'_i \leftarrow V(\tau'_i)$; \ $s'_i \leftarrow \ssc{Score}(\text{fb}'_i)$
    \State accept $\tau'_i$ w.p. $\min\!\left(1,\,(s'_i/s_i)^{1/T_{\text{smc}}}\right)$

  \EndFor
  \State $\mathcal{P}^{(t)} \leftarrow \ssc{UpdateParticles}(\mathcal{P}^{(t-1)},\{\tau'_i\})$
\EndFor
\State \Return $\mathcal{P}^{(T)}$
\end{algorithmic}
\end{algorithm}
\vspace{-10pt}

\paragraph{Scoring:} Let $V$ denote the verifier (\prm) and $\tau$ a candidate reasoning trace sampled from the generator $G$. Before scoring, we deterministically coerce $\tau$ into an explicit stepwise representation,\footnote{The normalizer first parses existing \texttt{<step>} blocks and, if none are found, splits on blank lines and wraps each segment in \texttt{<step>} tags.}, i.e., $\tau \leftarrow \ssc{StepwiseNormalize}(\tau)$. The verifier then generates step-level feedback $\text{fb}=V(\tau)$, which we map to a scalar consistency score $s(\tau)\in[0,1]$ using a deterministic rule $\textsc{Score}(\cdot)$, where higher values indicate stronger internal coherence (see Appendix~\ref{appendix:scoring}). We then convert this score into an unnormalized importance weight
\begin{equation}
w(\tau) \propto s(\tau)^{1/T_{\text{smc}}},
\end{equation}
where $T_{\text{smc}}$ controls the exploration-exploitation tradeoff (smc stands for Sequential Monte Carlo). Lower temperatures concentrate probability mass on high-scoring candidates, while higher temperatures preserve diversity.

\vspace{-5pt}
\paragraph{Resampling:} Importance weighting can concentrate probability mass on a small subset of candidates, which we quantify by computing the \emph{effective sample size (ESS)}
\begin{equation}
\mathrm{ESS} = \frac{\left(\sum_i w_i\right)^2}{\sum_i w_i^2}.
\end{equation}
If $\mathrm{ESS}$ falls below a threshold (e.g., $\alpha N$),\footnote{ESS ranges from 1 to the population size $N$. A high ESS indicates relatively uniform weights and preserved diversity, while a low ESS indicates collapse, where a small number of trajectories dominate the population.} we resample: high-weight candidates are duplicated and low-weight candidates discarded.
Resampling reallocates compute toward promising trajectories while maintaining population-level stability.

\vspace{-5pt}
\paragraph{Stochastic local refinement:} Resampling allocates more computation to promising candidates but does not introduce new reasoning directions. To enable correction and exploration, \method applies a stochastic local refinement to each candidate.

Given a trace $\tau$, an iterator model $I$ proposes a refined trace using \prm feedback. Most proposals perform local refinements to the current reasoning, while a small fraction (controlled by a hyperparameter $\eta$) explore alternative reasoning directions (see Appendix~\ref{appendix:stochastic_refinement}). Define the acceptance ratio
\begin{equation}
r_w = \frac{w(\tau')}{w(\tau)} = \left(\frac{s(\tau')}{s(\tau)}\right)^{1/T_{\text{smc}}},
\label{eq:acceptance_main}
\end{equation}

The proposal is accepted with probability $A(\tau\rightarrow\tau') = \min(1,r_w)$,
which favors increases in \prm score while still allowing score-decreasing moves with some probability. This turns refinement into a score-guided local search that balances correction and exploration.\footnote{This acceptance rule is inspired by Metropolis-style stochastic search. It is not an exact Metropolis-Hastings correction, as the \fssc{LLM} proposal density is intractable.}

\vspace{-5pt}
\paragraph{Practical safeguards:} We incorporate two stabilizers to prevent pathological population dynamics: (i) \textbf{conflict arbitration} resolves cases where distinct answers receive similarly high \prm scores by using a comparator model and clamping conflicting candidates to a minimum score $c$ (See Appendix~\ref{appendix:arbitration}); and (ii) \textbf{clone capping} limits the fraction $k$ of the population that can be occupied by duplicated traces during resampling, which prevents pathological  collapse and keeps behavior invariant to population size.%

\vspace{-5pt}
\paragraph{\bsc{PRM}-guided solution aggregation:} After refinement, \method converts the final candidate population $P^{(T)}=\{\tau_i\}_{i=1}^{N}$ into a single prediction using \prm-score voting. For each unique extracted answer $a$, we  aggregate its \prm score
\begin{equation}
\label{eq:prm_vote}
S(a)=\sum_{i: Ans({\tau_i}) = a} s(\tau_i),
\end{equation}
where $Ans({\tau_i})$ denotes the answer extracted from trajectory
$\tau_i$. The final prediction is then $\hat{a}=\arg\max_a S(a)$. This aggregation favors answers supported by higher-quality reasoning rather than by frequency alone.%

\section{Experimental setup}
\label{sec:exp_setup}
Having established our \method method, we now evaluate its effectiveness as an end-to-end inference algorithm. Our goal is to measure how different \deepthink methods translate additional compute into accuracy. To enable a controlled comparison, all methods share the same inference configuration, including identical backbone models, population width, refinement depth, and initialization. Token usage may vary across methods and is reported explicitly. We fix population creation using Sample-N so that all methods operate on the same initial candidates. For solution aggregation, we report majority voting and \ssc{LLM}-based aggregation for all methods, while \method additionally supports \prm-score voting as part of its inference procedure. Below, we describe the baselines,  benchmarks, models, inference protocol, and evaluation metrics.

\vspace{-5pt}
\subsection{Baselines}
We compare \method with a diverse set of representative methods that differ in how they allocate compute across candidate solutions.

\vspace{-5pt}
\paragraph{Simple Voting (no refinement):} The final answer is  selected by majority vote or \prm-score vote over the initial candidate population without iterative refinement.

\vspace{-5pt}
\paragraph{SciMaster~\citep{chai2025scimaster}:} Each candidate reasoning trace is independently critiqued and rewritten at every iteration. This method focuses on local error correction but rewrites entire solutions without an explicit correctness signal.
\vspace{-5pt}
\paragraph{Agentic Debate~\citep{wang2025retrievalaugmented}:} A multi-agent framework in which each candidate revises itself using information from other candidates in the population, enabling peer-to-peer information flow. This collaborative refinement can improve consensus but may reinforce shared errors.

\vspace{-5pt}
\paragraph{MAD Conformist~\citep{choi2025debate}:} A majority-driven framework that anchors the population to the current majority answer. Candidates that agree with the majority are preserved, while non-conforming candidates are selectively refined toward consensus.

\vspace{-5pt}
\paragraph{MAD Follower~\citep{choi2025debate}:} A majority-driven variant that replaces a fraction of the population with candidates that explicitly adopt the majority solution before refinement. This more aggressively steers the population toward agreement.

\vspace{-5pt}
\paragraph{Recursive Self-Aggregation~\citep{venkatraman2025recursive}:} An aggregation-driven framework that repeatedly synthesizes subsets of candidate solutions into new aggregate solutions, which are then reintroduced into the population.

\vspace{-5pt}
\subsection{Evaluation benchmarks}
We evaluate on three reasoning benchmarks: AIME25 and HMMT25 from MathArena~\citep{balunovic2025matharena}, %
and GPQA Diamond~\citep{rein2024gpqa}. Since \deepthink inference is compute-intensive, we cap GPQA Diamond at the first 120 examples to keep evaluation tractable. AIME25 and HMMT25 are evaluated in full. For \method and Recursive Self-Aggregation, we additionally evaluate on code generation using LiveCodeBench v5 and v6~\citep{jain2024livecodebenchholisticcontaminationfree}; results are reported in Section~\ref{sec:coding_generation}.

\vspace{-5pt}
\subsection{Models}
We evaluate nine generator models across two families. We use \texttt{gpt-oss-20b} as the primary backbone for \method and other \deepthink methods, while \texttt{gpt-oss-120b} serves mainly as a strong zero-shot baseline. We also experiment with the Qwen3 family~\citep{yang2025qwen3}, including Qwen-1.7B, Qwen-4B, Qwen-14B, and Qwen-30B-A3B (hybrid pretrained + post-trained variants). We further evaluate three additional Qwen-4B variants (Base, Instruct, and Thinking) to examine how \method interacts with models that differ in training objective and reasoning specialization.

\vspace{-2pt}
Unless otherwise specified, the generator $G$, verifier $V$, iterator $I$, and comparator $C$ models are all instantiated from the same backbone model (Appendices~\ref{sec:appendix_verifier},~\ref{sec:appendix_proposals},~\ref{sec:appendix_arbitration} contain the prompts). One exception is our cross generator-verifier scaling experiment (Section~\ref{sec:exp_qwen}), where the verifier is allowed to differ in size from the generator to study the effect of verifier strength. For the generator and iterator, which require diverse responses for \deepthink, we use stochastic decoding with temperature $=0.8$ and $top\_p=0.9$. For the verifier and comparator, we always use deterministic decoding (temperature $=0$).

\vspace{-5pt}
\subsection{Inference protocol}
For each problem, we sample from the generator to initialize a population of $N=10$ candidate reasoning traces. These traces form the shared starting point for all methods. Each method then performs $T=5$ refinement iterations under the same inference configuration while maintaining a fixed-width population. For \method, we use temperature $T_{\text{smc}}=0.8$, ESS threshold $\alpha=0.5$, noise $\eta=0.1$, clamp $c = 0.3$, clone cap $\kappa = 0.3$. All baselines are implemented within our codebase for consistency, following their official source code and configuration details (if any) as closely as possible. \texttt{gpt-oss} models use the medium reasoning effort setting, while Qwen models are limited by a 32K token context window by default.

\vspace{-5pt}
\subsection{Evaluation metrics}
In addition to final accuracy, we evaluate inference behavior using population-level and compute-aware metrics.

\vspace{-5pt}
\paragraph{Population accuracy:} At each refinement depth $t$, each candidate $\tau^{(t)}_{q,i}$ yields an extracted answer $Ans(\tau^{(t)}_{q,i})$. We measure the fraction of correct candidates:
\begin{equation}
\label{eq:popacc}
\mathrm{PopAcc}(t)=\mathbb{E}_{q\sim\mathcal{D}}
\left[\frac{1}{N}\sum_{i=1}^{N}\mathbf{1}\{Ans(\tau^{(t)}_{q,i})=y_q\}\right],
\end{equation}
which captures whether refinement increases the density of correct candidates.

\vspace{-5pt}
\paragraph{NetFlip:}
To separate genuine correction from ``random-walk'' rewriting, we measure directional correctness change between consecutive refinement steps. Let $I2C(t)$ be the number of incorrect$\rightarrow$correct transitions and $C2I(t)$ the number of correct$\rightarrow$incorrect transitions from step $t{-}1$ to $t$. We define
\begin{equation}
\mathrm{NetFlip} = \sum_{t \in T} \left( I2C(t) - C2I(t) \right)
\end{equation}
Positive values indicate net error correction, while negative values indicate net degradation.
\vspace{-5pt}
\paragraph{Compute:} We report compute as total token usage across all model calls during inference (including generation, verification, and comparison) and analyze accuracy–compute tradeoffs across methods. We also report estimated cost for relative comparison under a fixed pricing assumption (see Table~\ref{tab:final_accuracy}).

\section{Results and discussion}
\label{sec:exp_main_results}
\begingroup
\renewcommand{\arraystretch}{1.5}
\newcommand{\pmstd}[1]{$_{\pm#1}$}
\begin{table*}[t]
\centering
\caption{\textbf{Main results across reasoning benchmarks.} \method is competitive with or outperforms strong \deepthink baselines across the evaluated settings.\protect\footnotemark\ See Section~\ref{sec:coding_generation} for results on coding generation benchmarks.
}
\label{tab:final_accuracy}

\begin{adjustbox}{max width=0.99\textwidth}
\begin{tabular}{lccccc}
\toprule
\textbf{Method} & \textbf{Aggregation} & \textbf{AIME25} & \textbf{HMMT25} & \textbf{GPQA Diamond} & \textbf{Cost (\$)} \\
\hline
\rowcolor{tealbg}\multicolumn{6}{l}{\textit{Non-\deepthink baselines}} \\ \hline

\texttt{gpt-oss-120b} (zero-shot) & -- & 77.8\pmstd{5.1} & 68.9\pmstd{8.4} & 69.7\pmstd{1.7} & 0.20 \\
\texttt{gpt-oss-20b} (zero-shot) & -- & 58.9\pmstd{10.7} & 44.4\pmstd{13.5} & 59.7\pmstd{2.5} & \textbf{0.08} \\
Majority Vote & Majority Vote & 76.7\pmstd{0.0} & 58.9\pmstd{1.9} & 65.8\pmstd{0.8} & 0.87 \\
\hline
\rowcolor{tealbg} \multicolumn{6}{l}{\textit{\deepthink baselines + Majority Vote}} \\ \hline
SciMaster & Majority Vote & 84.4\pmstd{1.9} & 70.0\pmstd{6.7} & 63.4\pmstd{2.0} & 6.60 \\
Agentic Debate & Majority Vote & 85.6\pmstd{1.9} & 72.2\pmstd{5.1} & 65.0\pmstd{0.8} & 4.67 \\
MAD Conformist & Majority Vote & 81.1\pmstd{1.9} & 66.7\pmstd{0.0} & 66.9\pmstd{1.3} & 1.71 \\
MAD Follower & Majority Vote & 80.0\pmstd{0.0} & 62.2\pmstd{1.9} & 64.7\pmstd{2.5} & 3.30 \\
Recursive Self-Aggregation & Majority Vote & 87.8\pmstd{1.9} & \textbf{77.8}\pmstd{1.9} & 68.6\pmstd{1.7} & 4.03 \\
\hline
\rowcolor{tealbg} \multicolumn{6}{l}{\textit{\deepthink baselines + \ssc{LLM} Aggregate}} \\ \hline
SciMaster & \ssc{LLM} Aggregate & 77.8\pmstd{6.9} & 70.0\pmstd{5.8} & 63.9\pmstd{4.1} & 6.87 \\
Agentic Debate & \ssc{LLM} Aggregate & 82.2\pmstd{5.1} & 75.6\pmstd{5.1} & 58.1\pmstd{1.3} & 4.80 \\
MAD Conformist & \ssc{LLM} Aggregate & 82.2\pmstd{3.8} & 65.6\pmstd{13.5} & 63.6\pmstd{1.3} & 1.81 \\
MAD Follower & \ssc{LLM} Aggregate & 83.3\pmstd{3.3} & 73.3\pmstd{3.3} & 63.3\pmstd{1.7} & 3.40 \\
Recursive Self-Aggregation & \ssc{LLM} Aggregate & 84.4\pmstd{1.9} & 74.4\pmstd{6.9} & 64.7\pmstd{3.5} & 4.15 \\
\hline
\rowcolor{tealbg} \multicolumn{6}{l}{\textit{\deepthink baselines + \prm-score Vote}} \\ \hline
SciMaster & \prm-score Vote & 83.3\pmstd{0.0} & 67.8\pmstd{1.9} & 64.2\pmstd{2.2} & -- \\
Agentic Debate & \prm-score Vote & 85.6\pmstd{1.9} & 75.3\pmstd{7.3} & 63.3\pmstd{1.4} & -- \\
MAD Conformist & \prm-score Vote & 82.2\pmstd{1.9} & 67.8\pmstd{1.9} & 66.9\pmstd{1.3} & -- \\
MAD Follower & \prm-score Vote & 81.1\pmstd{1.9} & 64.4\pmstd{5.1} & 63.6\pmstd{2.5} & -- \\
Recursive Self-Aggregation & \prm-score Vote & 87.8\pmstd{1.9} & \textbf{77.8}\pmstd{1.9} & 68.9\pmstd{2.5} & -- \\
\hline
\rowcolor{tealbg} \multicolumn{6}{l}{\textit{Our methods}} \\ \hline
\prm-score Vote & \prm-score Vote & 77.8\pmstd{1.9} & 59.7\pmstd{0.5} & 64.7\pmstd{2.5} & 1.01 \\
\method{} & Majority Vote & 87.8\pmstd{1.9} & 72.2\pmstd{1.9} & \textbf{71.7}\pmstd{2.5} & 6.76 \\
\method{} & \ssc{LLM} Aggregate & 86.7\pmstd{5.8} & 75.6\pmstd{3.8} & 68.3\pmstd{2.2} & 7.11 \\
\method{} & \prm-score Vote & \textbf{90.0}\pmstd{3.3} & 75.4\pmstd{4.0} & 71.4\pmstd{1.3} & 6.76 \\
\bottomrule
\end{tabular}
\end{adjustbox}
\end{table*}
\footnotetext{Estimated cost is computed from the total token usage across all datasets, assuming a 60/40 input/output split. Input and output tokens are priced at \$0.05 and \$0.20 per million tokens (\url{https://www.together.ai/}). Values are shown for relative comparison, since absolute cost depends on the provider and the input/output ratio.\\
See Section~\ref{sec:exp_qwen} for our experiments with Qwen3.
}
\endgroup

\begin{figure*}[t]
    \centering
    \includegraphics[width=\linewidth]{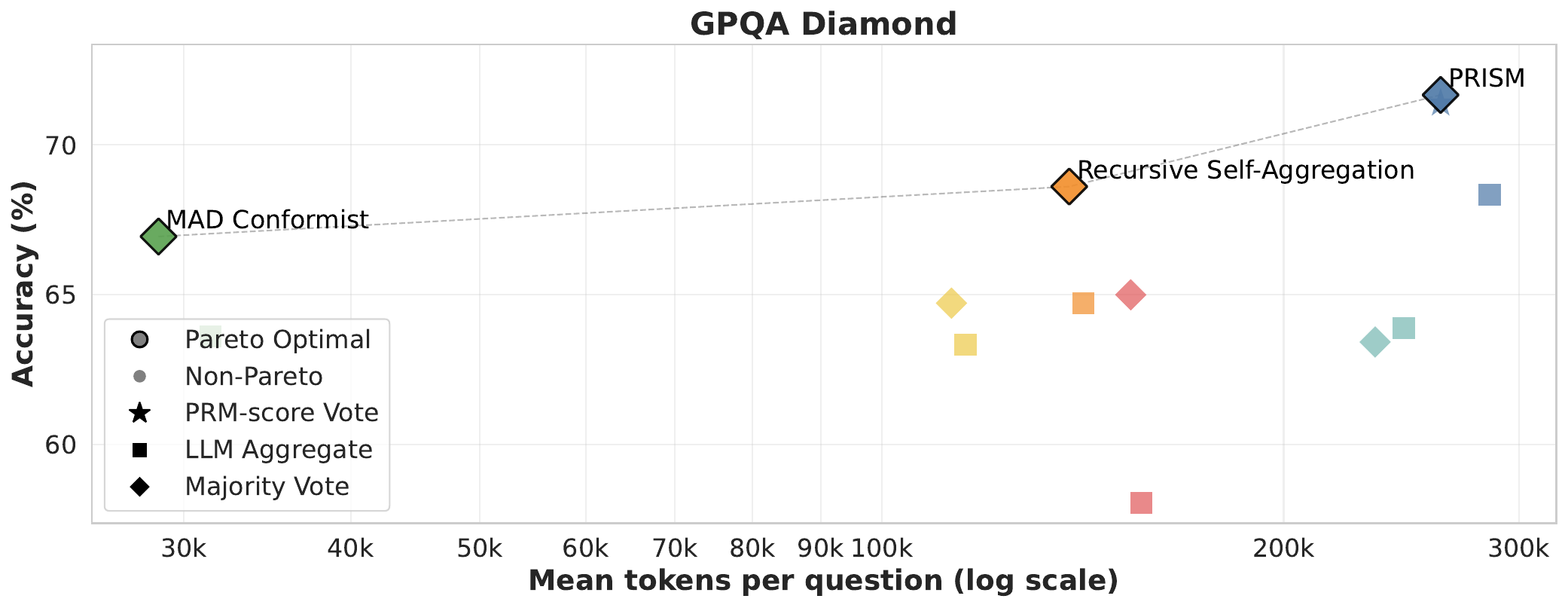}
    \caption{\textbf{Compute–accuracy tradeoff on GPQA Diamond}. Each point represents a method configuration (enhancement + aggregation). The x-axis shows mean token usage per question, and the y-axis shows final accuracy. \method achieves the highest accuracy among the evaluated methods.}
    \label{fig:pareto_baselines}
\end{figure*}

\begin{figure*}[h]
    \centering
    \includegraphics[width=0.8\linewidth]{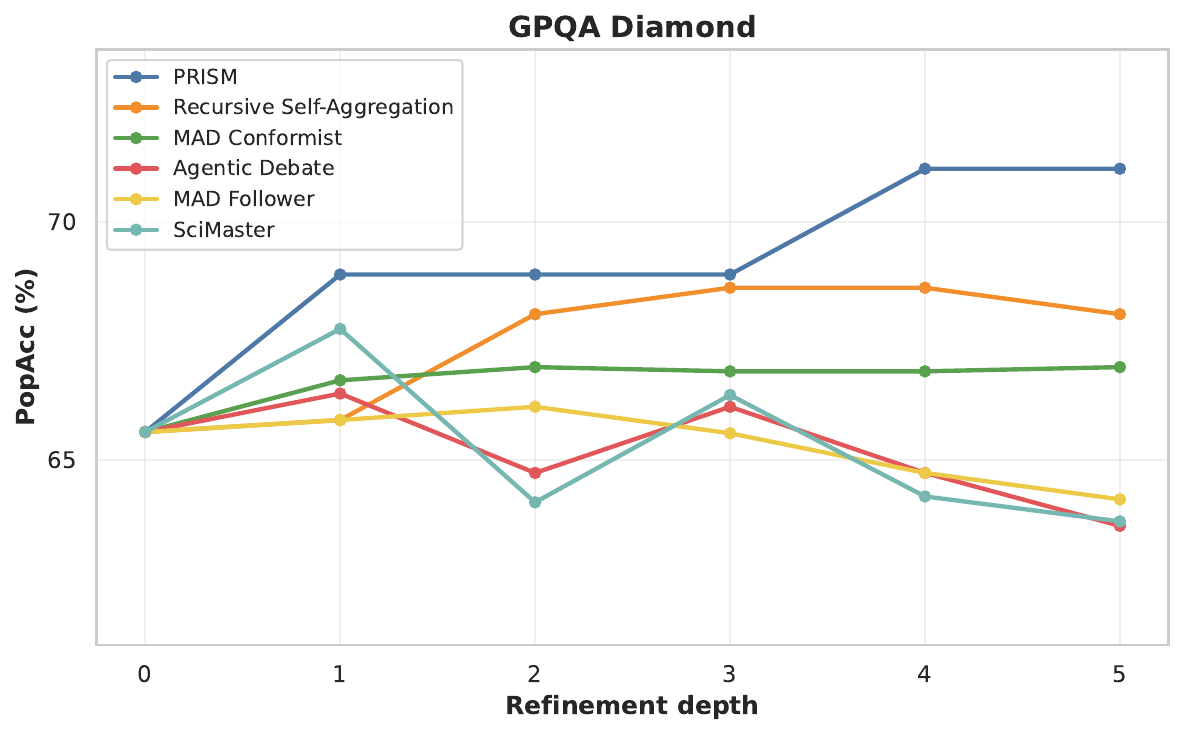}
    \caption{\textbf{Population quality vs. refinement depth on GPQA Diamond.} Non-\prm-based methods often oscillate or degrade with depth, whereas \method exhibits stable upward population dynamics.}
    \label{fig:popacc_depth_baselines}
\end{figure*}

\begin{figure*}[t]
    \centering
    \includegraphics[width=0.85\linewidth]{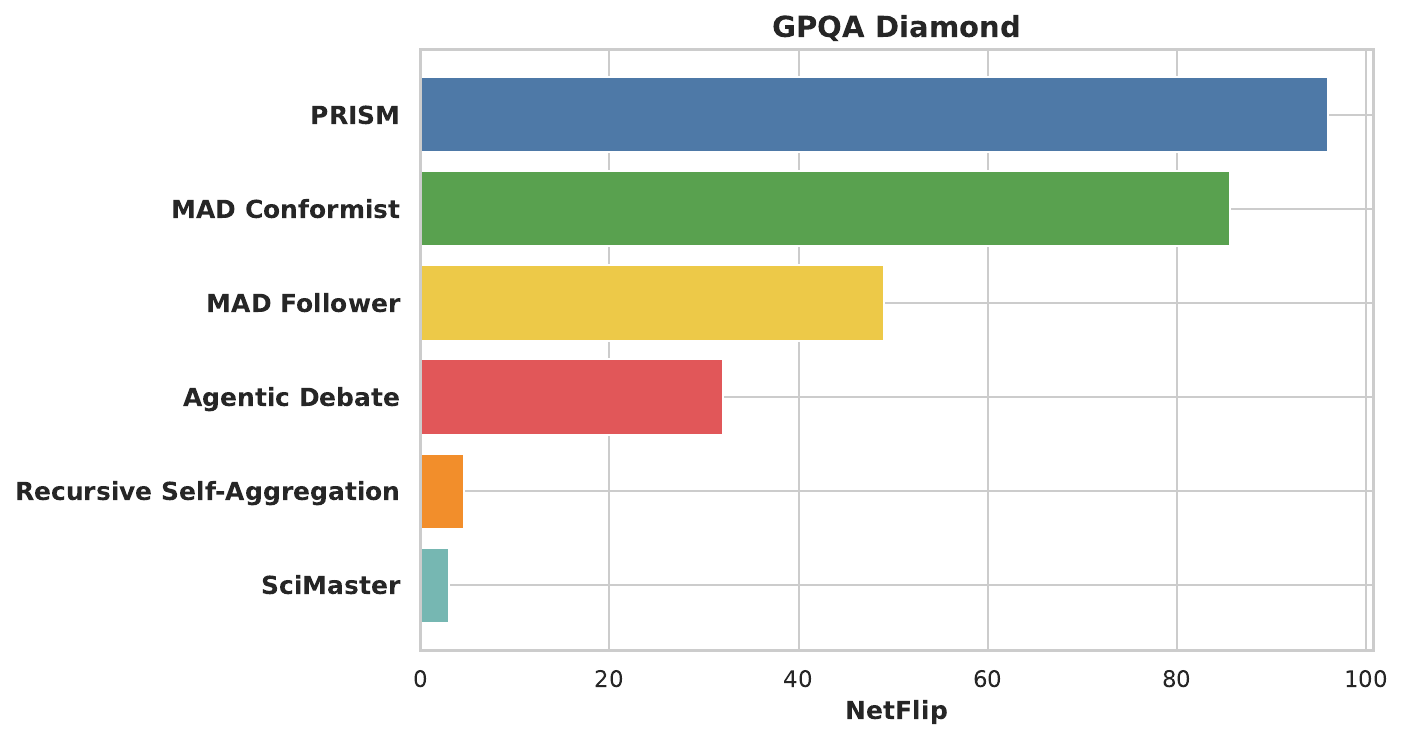}
    \caption{\textbf{Directional correction across enhancement depth on GPQA Diamond.} Bars show NetFlip aggregated over depth and averaged across questions. Positive values indicate more incorrect$\rightarrow$correct than correct$\rightarrow$incorrect transitions, reflecting genuine correction rather than ``random-walk'' reshuffling.}
    \label{fig:netflip_baselines}
\end{figure*}
\vspace{-3mm}

\begin{figure}[h]
    \centering
    \includegraphics[width=\linewidth]{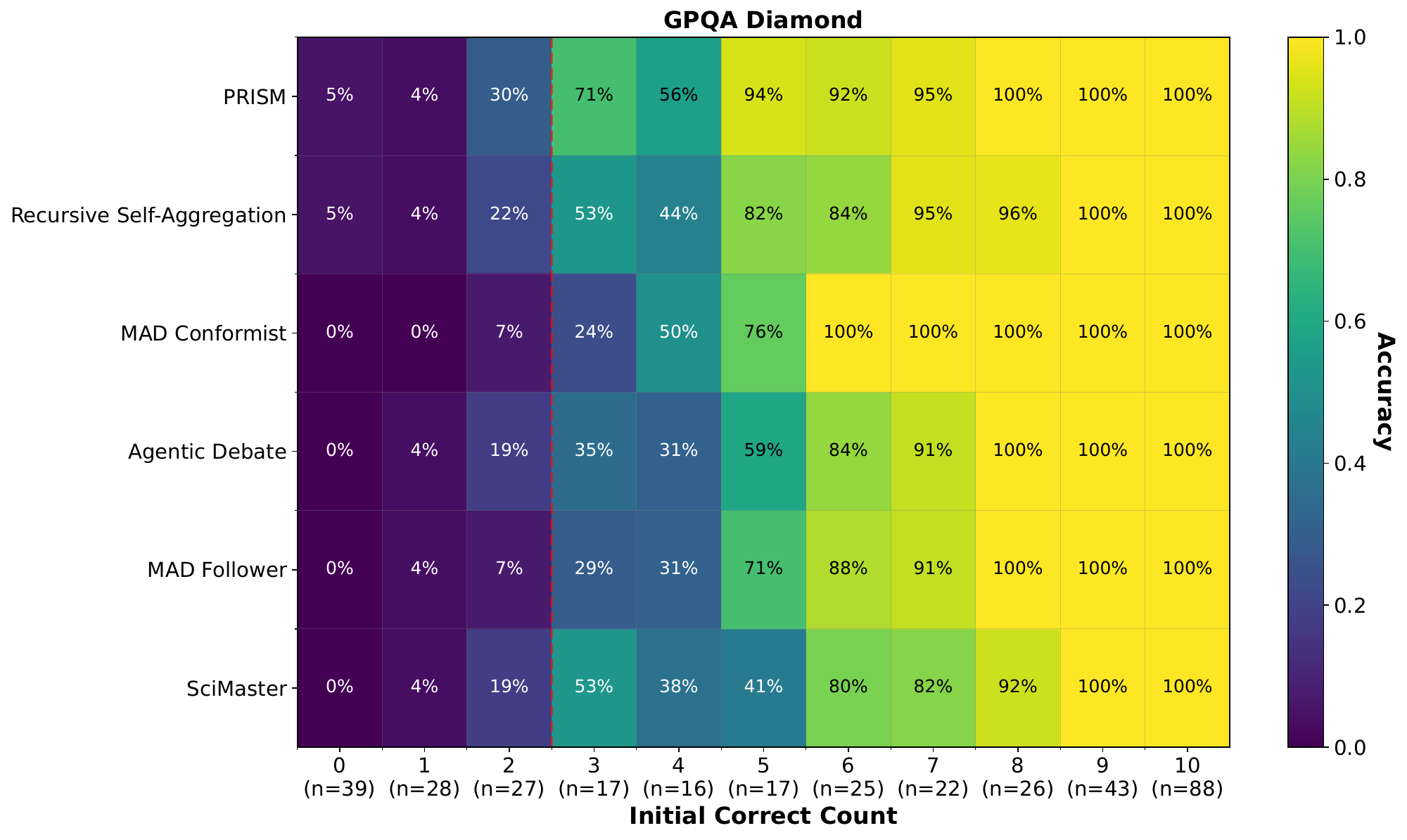}
    \caption{\textbf{GPQA performance conditioned on initial population quality.} Each column groups problems by how many correct solutions appear in the initial candidate pool, and each cell reports the final accuracy after refinement and majority voting. While most methods struggle when few correct candidates are present at the start, \method maintains substantially higher accuracy in these low-correctness regimes, which shows stronger resilience to weak initial populations.}
    \label{fig:gpqa_low_signal_heatmap}
\end{figure}

We evaluate \method under controlled settings and compare it with state-of-the-art \deepthink baselines. \method consistently improves accuracy, often enabling a 20B model to exceed a 120B zero-shot baseline. \method also yields more stable refinement dynamics, with consistent directional error correction and improved robustness under weak or noisy initial populations.

\vspace{-5pt}
\subsection{Main results}
\vspace{-2pt}
\paragraph{\bssc{PRISM} consistently improves final accuracy across the evaluated benchmarks:} Table~\ref{tab:final_accuracy} reports final accuracy as well as the standard deviation using the same seed population and backbone \ssc{LLM} (\texttt{gpt-oss-20b}). The result is averaged across three runs. Across datasets, \method is competitive with or exceeds strong \deepthink baselines under the evaluated settings.

\vspace{-2pt}
On AIME25, \method with \prm-score voting achieves 90.0\%, surpassing Recursive Self-Aggregation (87.8\%) and Agentic Debate (85.6\%). On HMMT, \method reaches 75.4\%, remaining competitive with debate-based and aggregation-heavy methods. On GPQA Diamond, \method attains 71.4\%, improving over Recursive Self-Aggregation (68.6\%) and all majority-driven approaches. Notably, with additional inference-time refinement, \method enables \texttt{gpt-oss-20b} to exceed the zero-shot accuracy of \texttt{gpt-oss-120b}.

\vspace{-2pt}
Figure~\ref{fig:pareto_baselines} shows the compute-accuracy tradeoff on GPQA. Across the evaluated methods, \method achieves the highest accuracy, demonstrating strong performance as additional inference-time computation is allocated to refinement.

\vspace{-2pt}
A key observation is that Majority Vote is already a strong baseline, reaching 65.8\% on GPQA. Several refinement-heavy methods fail to surpass it despite higher token usage, suggesting that iterative rewriting alone does not reliably improve the candidate population. In contrast, \method uses step-level verification to guide refinement toward higher-quality reasoning, resulting in stronger final accuracy.

\vspace{-5pt}
\subsection{Diagnosing refinement dynamics}
\label{sec:exp_ablations}
We next analyze population-level dynamics to understand why additional refinement compute often fails to translate into correctness.

\vspace{-5pt}
\paragraph{\bssc{PRISM} makes refinement directionally corrective:}
Figure~\ref{fig:popacc_depth_baselines} shows population accuracy as a function of refinement depth. Across most non-\prm-based methods, population quality oscillates or degrades rather than improving monotonically, which indicates that refinement updates behave largely as stochastic perturbations rather than directional improvements. In contrast, \method shows stable upward population dynamics, which suggests that \prm-guided acceptance filtering suppresses harmful updates and preserves correct trajectories.

\vspace{-2pt}
To assess whether refinement corrects more errors than it introduces, we measure NetFlip. On GPQA Diamond, most baselines exhibit small positive net flip, which suggests limited directional correction (Figure~\ref{fig:netflip_baselines}). In many cases, correct trajectories are degraded nearly as often as incorrect ones are repaired. In contrast, \method produces substantially larger positive NetFlip values, which shows that refinement is more consistently aligned with correctness.

\vspace{-2pt}
\paragraph{\method preserves correct minority solutions and maintains competitive accuracy in low-correctness regimes:}
To examine how different methods behave under varying initial population quality, we group instances by the number of correct candidates present in the initial population before refinement and report final accuracy within each group using majority voting for all methods. We visualize our results on GPQA Diamond in Figure~\ref{fig:gpqa_low_signal_heatmap}.

\vspace{-2pt}
Majority-driven refinement (e.g., MAD Conformist/Follower) stabilizes populations by anchoring candidates to frequent solutions, but suppresses correct minority trajectories in favor of more frequent but incorrect reasoning. When the initial majority is wrong, this drives convergence to a confident but incorrect answer, especially in low-correctness regimes where recovery is rare.

\vspace{-2pt}
In contrast, \method degrades substantially less and achieves higher accuracy when the initial population contains few correct candidates, which indicates that it can ``bootstrap'' from weak populations rather than amplifying dominant trajectories. As refinement is guided by step-level correctness signals rather than frequency, correct minority trajectories are preserved, avoiding majority dilution and enabling recovery.

\vspace{-5pt}
\paragraph{Directional refinement makes aggregation more reliable:} Aggregation quality depends on the refined population. For most baselines, switching from majority vote to \ssc{LLM} aggregation often reduces accuracy, which suggests that the aggregator can rationalize confident but incorrect traces when the population is noisy (see Table~\ref{tab:final_accuracy}). In contrast, \method maintains stable performance under \ssc{LLM} aggregation, indicating that directional refinement produces more accurate and internally consistent populations. \prm-score voting further mitigates majority dilution when correct answers are rare (e.g., on AIME), while having little effect when refinement already promotes correct trajectories into the majority (e.g., on GPQA).

\vspace{-5pt}
\paragraph{Directional refinement and population stabilization in \bssc{PRISM}:} We now examine \method's internal refinement dynamics. At each iteration $t$, \method assigns \prm-based weights to candidate traces and triggers resampling when weight concentration becomes severe, which is tracked by normalized $ESS/N$ (higher indicates greater population diversity, i.e., more evenly distributed weights). As shown in Table~\ref{tab:resampling_clonecap_diagnostics}, early refinement exhibits strong weight concentration ($ESS/N\approx$ 0.24-0.33), which triggers frequent resampling (50.3\%–79.5\%). By $t=4$, $ESS/N$ rises to 0.81-0.88 and resampling drops to 3.3\%–8.5\%, indicating stabilization of the candidate pool. Without clone capping, a single trajectory would often exceed a 30\% share of the population ($P(max > 0.3)$) and full takeover ($P(max = 1)$) would occur non-trivially. The clone cap prevents such collapse and preserves diversity needed for reliable aggregation and recovery.

\begin{table}[t]
\centering
\caption{\textbf{Resampling dynamics}. $t=0$ is initialization; $t=4$ is after four refinement steps. $ESS/N \in [0, 1]$ measures weight concentration (higher means more evenly distributed weights). Resampling rate is the fraction of iterations that triggered resampling. Dominance rate w/o clone cap (\%) reports the probability that a single candidate exceeds 30\% of the next generation ($P(max>0.3)$) and the probability of full takeover ($P(max=1)$).}
\label{tab:resampling_clonecap_diagnostics}
\begin{adjustbox}{max width=0.7\textwidth}
\begin{tabular}{lcc|cc|cc}
\toprule
\textbf{Dataset} &
\multicolumn{2}{c|}{\textbf{ESS/N}} &
\multicolumn{2}{c|}{\textbf{Resampling rate}} &
\multicolumn{2}{c}{\textbf{Dominance rate w/o clone cap}} \\
& \textbf{$t{=}0$} & \textbf{$t{=}4$} & \textbf{$t{=}0$} & \textbf{$t{=}4$} &
$P(\max{>}0.3)$ & $P(\max{=}1)$ \\
\midrule
AIME & 0.282 & 0.879 & 0.756 & 0.033 & 0.854 & 0.282 \\
HMMT & 0.237 & 0.812 & 0.795 & 0.085 & 0.768 & 0.134 \\
GPQA & 0.333 & 0.838 & 0.503 & 0.064 & 0.989 & 0.316 \\
\bottomrule
\end{tabular}
\end{adjustbox}
\vspace{2pt}
\end{table}

\begin{table}[t]
\centering
\caption{\textbf{Proposal acceptance dynamics}. $r$ is the \prm score ratio between the proposal and the current trace. $P(accept | r_s<1)$ reports the downhill acceptance rate (percentage of lower-scoring proposals that are accepted). $\mathbb{E}[(s(\tau') \mid \text{accepted}]$ and $\mathbb{E}[s(\tau') \mid \text{rejected}]$ denote the average \prm score of proposed traces when accepted or rejected, respectively. Values are mean across runs with min-max in parentheses.}
\label{tab:accept_reject_diagnostics}
\begin{adjustbox}{max width=0.7\textwidth}
\begin{tabular}{lccc}
\toprule
\textbf{Dataset} & $P(accept | r_{w}<1)$ & $\mathbb{E}[(s(\tau') \mid \text{accepted}]$ & $\mathbb{E}[s(\tau') \mid \text{rejected}]$ \\
\midrule
AIME & 0.101 (0.094-0.106) & 0.794 (0.781--0.805) & 0.025 (0.023--0.026) \\
HMMT & 0.179 (0.138-0.238) & 0.692 (0.677--0.713) & 0.045 (0.031--0.060) \\
GPQA & 0.104 (0.087-0.117) & 0.763 (0.756--0.778) & 0.023 (0.021--0.024) \\
\bottomrule
\end{tabular}
\end{adjustbox}
\end{table}

\vspace{-2pt}
After resampling, \method proposes stochastic updates to candidate solutions, and accepts or rejects each proposal based on the score ratio $r_s$. Table ~\ref{tab:accept_reject_diagnostics} shows that accepted proposals consistently receive higher \prm scores than rejected ones, and score-decreasing proposals (downhill moves) are still accepted with non-zero probability (0.101-0.179), which preserves exploration. The number of accepted proposals correlates positively with improvements in population correctness on AIME and GPQA ($r = 0.152$ and 0.172), while remaining weaker on HMMT ($r = 0.023$), suggesting that the \prm provides a useful, though imperfect, local signal, while \prm's acceptance rule still maintains exploration through occasional lower-scoring updates.

\vspace{-5pt}
\subsection{Experiments with Qwen3 model family}
\label{sec:exp_qwen}
\vspace{-5pt}
\paragraph{\bssc{PRISM} generalizes across models and benefits from stronger verifiers:} Across AIME, HMMT, and GPQA, \method consistently improves over the zero-shot baseline, with larger gains on smaller models, as shown in Figures~\ref{fig:qwen_model_size_aime}--\ref{fig:qwen_model_size_gpqa} in Appendix~\ref{sec:appendix_figures}. Figures~\ref{fig:qwen_pareto_aime}--\ref{fig:qwen_pareto_gpqa} provide the corresponding compute-accuracy comparison, showing the token usage associated with these performance gains. We further probe the role of verifier strength by pairing Qwen3 generators and verifiers of different sizes. Performance improves further with stronger verifiers (e.g., pairing a 1.7B generator with 14B or 30B verifiers; Figure~\ref{fig:qwen_cross_gen_ver}). Finally, to assess robustness across model variants, we apply \method to Qwen-4B Base, Instruct, and Thinking. Across datasets, \method produces larger gains for weaker variants (e.g., base) and substantially narrows the gap to stronger models (e.g., thinking).

\subsection{Evaluation on code generation}
\vspace{-5pt}
\label{sec:coding_generation}
\begin{table}[t]
\centering
\caption{\textbf{\method consistently outperforms Recursive Self-Aggregation (RSA) on code generation benchmarks.} Average accuracy on LiveCodeBench v5 and v6 under the same setup as the main experiments.}
\begin{tabular}{lcc}
\toprule
Benchmark & RSA & \method \\
\midrule
LiveCodeBench v5 & 74.2 & \textbf{80.3}  (+6.1) \\
LiveCodeBench v6 & 71.1 & \textbf{72.8}  (+1.7) \\
\bottomrule
\end{tabular}
\label{tab:coding_results}
\vspace{-5pt}
\end{table}

While our primary focus is on mathematical and scientific reasoning, we also evaluate \method on code generation using the LiveCodeBench v5 and v6 benchmarks~\citep{jain2024livecodebenchholisticcontaminationfree}, following the same inference protocol as in our main experiments. Our results are shown in Table~\ref{tab:coding_results}. \method consistently improves over Recursive Self-Aggregation, achieving an average accuracy of 80.3\% on v5 (+6.1\%) and 72.8\% on v6 (+1.7\%). These results indicate that \prm-guided refinement is not specific to our target domain, but instead provides a general mechanism for improving structured reasoning, including in code generation. Additional implementation and evaluation details are provided in Appendix~\ref{appendix:coding_benchmark}.

\vspace{-5pt}
\section{Related work}
\paragraph{Test-time scaling and \bssc{DeepThink} frameworks:} A growing line of work studies how additional inference-time compute can trade off with pretraining compute to improve reasoning performance. Broadly, these approaches scale reasoning along two axes. \textbf{\emph{Sequential scaling}} generates solution attempts iteratively, where later attempts condition on earlier ones so that the model extends and refines its reasoning~\citep{snell2025scaling,muennighoff-etal-2025-s1}. For example, ~\citet{muennighoff-etal-2025-s1} rely on budget forcing, which artificially truncates or extends reasoning chains to control compute allocation. In contrast, \textbf{\emph{parallel scaling}} generates multiple solution attempts concurrently and selects among them, for example, by using majority voting or Best-of-N strategies~\citep{snell2025scaling,brown2025large}. Hybrid approaches combine sequential reasoning with parallel exploration, which includes tree-based inference methods such as Monte Carlo Tree Search~\citep{zhang2023planning,zhou2024language} and guided beam search~\citep{xie2023selfevaluation}, as well as approaches that incorporate process-level rewards to guide sequential search~\citep{wu2025inference}.

Our work situates itself within the emerging class of \deepthink frameworks, which extends the parallel paradigm by synergistically refining and combining interdependent reasoning paths rather than treating candidates as independent samples~\citep{gemini25deepthink,gemini3deepthink,fu2026deep,dong2026generalized}. Methods such as SciMaster~\citep{chai2025scimaster} and Recursive Self-Aggregation~\citep{venkatraman2025recursive} rely on iterative refinement and structured aggregation to progressively improve candidate solutions. In contrast, approaches like Agentic Debate~\citep{wang2025retrievalaugmented} and MAD Conformist/Follower~\citep{choi2025debate} emphasize agentic interaction and majority-driven consensus to coordinate reasoning across candidates.

\vspace{-5pt}
\paragraph{Process reward modeling:} Process Reward Models (\ssc{PRMs}) provide step-level supervision over reasoning steps~\citep{lightman2024lets}. Prior work uses process rewards for ranking, filtering, and training-time optimization~\citep{wang-etal-2024-math,wu2025inference}. Recent work also incorporates process rewards into inference-time search. \citet{setlur2025rewarding} develop progress-based process verifiers for test-time search and reinforcement learning, while~\citet{park2025ensembling} use a process reward model to guide tree search over intermediate reasoning steps. \method focuses on a different setting: population refinement in \deepthink systems. It uses step-level verification to guide population resampling, stochastic local refinement, and final aggregation.

\vspace{-5pt}
\section{Conclusion}
We introduced \method, a \prm-guided inference framework that improves the refinement of candidate solutions in \deepthink systems. At a high level, \method uses step-level feedback to guide population resampling and stochastic local refinement. This process directs refinement toward higher-quality reasoning while preserving diversity. Across challenging mathematical and scientific reasoning benchmarks, \method is competitive with or outperforms strong \deepthink baselines. Our analysis further shows consistent net-directional correction during refinement and strong performance when the initial population contains few correct candidates. These results suggest that reliable step-level verification is a key ingredient for improving inference-time reasoning.

\section{Limitations}
Our conclusions should be viewed in light of several design choices that constrain the scope of this study.
\paragraph{PRM instantiation: } \method depends on a step-level scalar signal to shape refinement and aggregation. In our experiments, the \prm is instantiated as a prompted model derived from the same base \ssc{LLM} and executed deterministically. This setup allows us to isolate the effect of \prm-guided population dynamics in a controlled setting, but it represents only one possible realization of the framework. In domains where stronger or externally grounded feedback is available, such as executable tests or formal verification tools, the scoring signal could be substantially more reliable. Therefore, our results demonstrate the effectiveness of integrating step-level signals into inference, but they do not fully characterize performance under richer reward sources.

\vspace{-5pt}
\paragraph{Step segmentation: } \method assumes that reasoning can be decomposed into meaningful steps. When segmentation does not align with coherent logical units, step-level scoring may be less informative, which weakens refinement guidance. Improved segmentation or more structured reasoning representations could further enhance the effectiveness of step-level signals.

\newpage
\bibliographystyle{plainnat}
\bibliography{references}

\newpage
\appendix
\appendix

\section{Step verifier and score construction}
\label{sec:appendix_verifier}

\subsection{Step verifier prompt}
\label{appendix:prm_mcmc_verifier_prompt}

We instantiate the verifier $V$ as a prompted auditor model that produces step-level
judgments in $\{+1,0,-1\}$ plus a \texttt{FINAL ANSWER CHECK}. The verifier is run
deterministically (temperature $0$) to reduce variance across
refinement iterations.

\paragraph{Input formatting:} Given a candidate trace $\tau$, we segment the reasoning into an ordered sequence of steps and wrap
each step in a \texttt{<step i="k">...</step>} tag (1-indexed). The verifier input consists of:
(i) the problem statement, (ii) the tagged step sequence, and (iii) the proposed final answer.

\paragraph{Output schema:} The verifier must return exactly one line per input step (in the same order), each containing a brief
note and a trailing score token in $\{+1,0,-1\}$, followed by a \texttt{FINAL ANSWER CHECK} line with a token in $\{+1,0,-1\}$. We parse only the score tokens and ignore free-form notes.

\begin{lstlisting}[style=llmstyle]
[SYSTEM_PROMPT]
    You are a strict auditor. Find the earliest mistake.
    Input: Problem, Steps (<step>...</step>), Proposed Answer.
    RULES:
    1) Judge EACH step independently (logic + math + consistency with Problem).
    2) If a step is wrong, mark it -1. Any later step that depends on it is -1.
    3) If a step is correct but trivial/doesn't advance the solution, mark 0.
    4) If anything is missing/ambiguous/unverifiable, mark -1.
    5) Be skeptical of 'clean' answers when derivation is not fully justified.
    OUTPUT (STRICT):
    Return EXACTLY one line per input step, in the SAME ORDER.
    Do NOT copy the original step text.
    Each line MUST be:
        <step i=\"k\">note <= 12 words ... +1|0|-1</step>
    The score token MUST be inside the <step ...> tag at the end.
    Do NOT put anything after </step>.
    Do NOT output the literal string '+1|0|-1' (choose exactly one).
    Example OK:
        <step i=\"1\">Algebra correct +1</step>
    Example NOT OK:
        <step i=\"1\">Algebra correct</step> +1
        <step i=\"1\">Algebra correct +1|0|-1</step>
    Use i=\"1\" for the first input step, i=\"2\" for the second, etc.
    After these, output exactly one final line:
        <step>FINAL ANSWER CHECK: +1|0|-1</step>
    FINAL ANSWER CHECK = -1 if Proposed Answer is wrong or not the requested quantity.
    FINAL ANSWER CHECK = 0 if cannot verify but might be correct.
    FINAL ANSWER CHECK = +1 only if it is clearly correct.
    Then output exactly one line:
        <answer>+1</answer> iff (all steps are +1) AND (FINAL ANSWER CHECK is +1).
        <answer>-1</answer> otherwise.
[USER_PROMPT]
    Problem: {question}
    Reasoning: {answer.reasoning}
    Proposed Answer: {answer.choice}
    Follow the system instructions exactly. Output only <step> lines and the final <answer> line.
\end{lstlisting}

\subsection{Parsing and failure handling}
\label{appendix:verifier_parsing}

Verifier outputs are treated as structured text. A verifier call is considered successful if it:
(i) emits one parsable score token for every non-final step,
(ii) uses only tokens in $\{+1,0,-1\}$, and
(iii) emits a parsable \texttt{FINAL ANSWER CHECK} token.

Any violation (e.g., missing lines, malformed tags, out-of-range tokens, or parsing-breaking extra text)
is treated as \texttt{VERIFICATION\_FAILED}, and the trace receives the fixed fallback score specified in
Section~\ref{appendix:scoring}. (We clamp scores to a small $\epsilon>0$ before forming ratios/weights to
avoid division-by-zero in acceptance computations.)

\subsection{Scoring, particle weights, and acceptance}
\label{appendix:scoring}

For each trace $\tau$, we compute a scalar verifier score $s(\tau)\in[0,1]$ from the
parsed feedback.

\paragraph{Step ratio:}
Let $n_{\text{correct}}, n_{\text{neutral}}, n_{\text{incorrect}}$ be the counts of $+1,0,-1$ labels over
non-final steps (excluding \texttt{FINAL ANSWER CHECK}). We define
\[
\text{step\_ratio:}
=\frac{n_{\text{correct}}+0.5\,n_{\text{neutral}}}
{n_{\text{correct}}+n_{\text{neutral}}+n_{\text{incorrect}}}.
\]
This weighted average treats correct steps as full credit, incorrect steps as no credit, and neutral
steps as partial credit. The partial weight mitigates penalties on long traces where neutral
labels are common, while still separating clean reasoning from clearly incorrect reasoning.

\paragraph{Final verdict:}
The \texttt{FINAL ANSWER CHECK} token yields the final verdict: \texttt{CORRECT} ($+1$),
\texttt{INCORRECT} ($-1$), or \texttt{NEUTRAL} ($0$). If the final line is missing but
step parsing succeeded, we mark it as \texttt{ MISSING}. If parsing fails, we mark \texttt{VERIFICATION\_FAILED}.

\paragraph{Scalar PRM score:}
The verifier provides two complementary signals: (i) \emph{how clean the reasoning is} (captured by
$\text{step\_ratio}$) and (ii) \emph{whether the final answer is actually correct} (captured by the final
verdict $v$). We combine them so that step-level quality can refine rankings \emph{within} a verdict class,
but cannot override the final correctness signal.

Concretely, we map $\text{step\_ratio}$ to a scalar score using a verdict-dependent affine transform
\[
s(\tau)=a_v+b_v\cdot \text{step\_ratio}.
\]
If $v=\texttt{INCORRECT}$, we cap the score below $0.3$ so a wrong final answer cannot achieve a high
score even if intermediate steps look plausible. If $v=\texttt{CORRECT}$, we floor the score at $0.5$
and scale upward with $\text{step\_ratio}$ so cleaner reasoning is preferred among correct solutions.
Neutral or missing final verdicts receive intermediate scaling to reflect uncertainty.

The parameters are:
\[
(a_v,b_v)=
\begin{cases}
(0,\;0.3), & v=\texttt{INCORRECT}\\
(0.5,\;0.5), & v=\texttt{CORRECT}\\
(0,\;0.6), & v=\texttt{NEUTRAL}\\
(0,\;0.8), & v=\texttt{MISSING}\\
(0,\;0), & v=\texttt{VERIFICATION\_FAILED}
\end{cases}
\]

\section{Stochastic refinement proposals}
\label{sec:appendix_proposals}

\subsection{Mixture proposal distribution}
\label{appendix:stochastic_refinement}

The iterator model $I$ proposes refinement moves $\tau\rightarrow\tau'$ using a two-component mixture:
\[
q(\tau' \mid \tau) \;=\; (1-\eta)\,q_{\text{local}}(\tau' \mid \tau,\text{\prm}(\tau))
\;+\; \eta\,q_{\text{explore}}(\tau' \mid x),
\]
where $x$ is the original problem instance and $\eta\in[0,1]$ controls exploration.

\paragraph{Local refinement ($q_{\text{local}}$):}
$I$ is conditioned on $x$, the current trace $\tau$, and structured \prm feedback on $\tau$.
The prompt instructs minimal edits: fix incorrect steps, preserve correct steps, and output a complete
revised trace with an explicit final answer.

\paragraph{Exploration ($q_{\text{explore}}$):}
$I$ is conditioned primarily on $x$ and is instructed to attempt a qualitatively different approach.
This component is intended to reduce mode collapse and introduce new reasoning directions.

\paragraph{Decoding and formatting constraints:}
Outputs must include an explicit final answer line for deterministic answer extraction. If $I$ produces an
unparsable output (e.g., missing final answer), we treat the proposal as a no-op ($\tau'=\tau$).

\subsection{Iterator prompt templates}
\label{appendix:iterator_prompts}
\begin{lstlisting}[style=llmstyle]
Problem:\n{state.question}

Previous reasoning:\n{prev_reasoning}
Previous choice: {particle.answer.choice}

Step-by-step verification:
{feedback_summary}

Refine the solution by fixing any -1 steps.
Maintain +1 steps unless you find a better approach.

Formatting requirements:

- Output ONLY <step>...</step> lines.
- Do NOT include markdown headings, separators, or text outside <step> tags.
- Keep the number of steps small and concise.
- Your LAST step MUST contain the final answer: 
<step>The final answer is: \\boxed{...}</step>
\end{lstlisting}
\begin{lstlisting}[style=llmstyle]
Problem:
{state.question}

Solve from scratch. Use a DIFFERENT approach than a standard solution if possible.

Formatting requirements:
- Output ONLY <step>...</step> lines.
- Keep steps concise.
- LAST step MUST be: <step>The final answer is: \\boxed{...}</step>
\end{lstlisting}

\section{Conflict arbitration}
\label{sec:appendix_arbitration}

PRM-style verifiers can assign similarly high scores to candidates with different extracted answers.
To avoid spending compute on incompatible modes, \method{} invokes an explicit arbitration step using the comparator
model $C$.

\subsection{Trigger, verdict parsing, and score clamping}
\label{appendix:arbitration}

\paragraph{Trigger:}
Let $\mathcal{A}$ be the set of distinct extracted answers in the population. We trigger arbitration when
(i) at least two answers exceed a near-perfect threshold, or (ii) the top answers are in a
near-tie.

\paragraph{Comparator output:}
$C$ outputs exactly one tag:
\texttt{<verdict>A</verdict>}, \texttt{<verdict>B</verdict>}, or \texttt{<verdict>NEITHER</verdict>}.
Malformed outputs are treated as \texttt{NEITHER}.

\paragraph{Score clamping:} Let $\tau_A,\tau_B$ be the compared traces and $c\in(0,1)$ be a clamp value. If the verdict is \texttt{A},
we clamp $s(\tau_B)\leftarrow \min(s(\tau_B),c)$ and keep $s(\tau_A)$ unchanged (symmetrically for \texttt{B});
for \texttt{NEITHER}, we clamp both. Since weights depend on scores ($w(\tau)=s(\tau)^{1/T}$), clamping reduces
the resampling probability of unresolved conflicts and limits their influence on subsequent refinement.

\subsection{Comparator prompt template}
\label{appendix:compare_prompt}
\begin{lstlisting}[style=llmstyle]
    You are a careful judge. Two solutions to the same problem both claim to be correct.
    Compare them and determine which one (if any) is actually correct.
    
    Problem:
    {problem}
    
    Solution A (answer: {choice_a}):\n{reasoning_a}
    
    Solution B (answer: {choice_b}):\n{reasoning_b}
    
    Instructions:
    1. Check each solution's reasoning and final answer.
    2. If Solution A is correct and B is wrong, output: <verdict>A</verdict>
    3. If Solution B is correct and A is wrong, output: <verdict>B</verdict>
    4. If BOTH are wrong or you cannot determine, output: <verdict>NEITHER</verdict>
    
    Output exactly one <verdict>...</verdict> tag with your decision.
\end{lstlisting}

\section{Coding generation benchmark details}
\label{appendix:coding_benchmark}
\subsection{Benchmark setup}
We additionally evaluate \method on code generation using two benchmark slices derived from LiveCodeBench. LiveCodeBench v5 contains problems with contest dates from October 1, 2024 to January 31, 2025, while LiveCodeBench v6 covers February 1, 2025 to April 30, 2025. In both cases, we exclude examples with invalid dates or incomplete test data, and report results on the first 120 filtered problems under a fixed deterministic ordering. No additional filtering by source or difficulty is applied. The resulting subsets contain a substantial fraction of challenging problems: LiveCodeBench v5 includes 28 easy, 31 medium, and 61 hard problems, while v6 contains 28 easy, 35 medium, and 57 hard problems. We report exact problem accuracy after reducing each method's final candidate population to a single program. A prediction is counted as correct only if the selected program passes the full evaluation suite.

\subsection{Evaluation protocol}
The coding experiments follow the same inference setup as in Sections~\ref{sec:exp_setup}. All methods share the same backbone model and inference configuration. We use \texttt{gpt-oss-20b} with a population width of 10, refinement depth of 5, and sampling temperature 0.8. For \method,
we adopt the same hyperparameters used in the main experiments, including $T_{\text{smc}} = 0.8$, ESS threshold $\alpha = 0.5$, and refinement noise $\eta = 0.1$.

\subsection{Code evaluation details}
Unlike the math and science benchmarks, correctness for coding tasks is determined by program execution rather than answer matching. Model outputs are converted into executable Python programs, normalized, and evaluated against the union of public and private test cases provided by the benchmark. A problem is considered solved only if all test cases pass. Public test cases may be used during refinement and final candidate selection, while private test cases are reserved strictly for final evaluation. Aggregation is therefore execution-aware. Recursive Self-Aggregation groups candidates based on behavioral equivalence over public tests rather than exact string matching.
In contrast, \method incorporates both behavioral agreement and \prm scores when selecting the final candidate, which enables it to prioritize solutions that are both functionally correct and supported by higher-quality reasoning.

\FloatBarrier
\section{Additional figures}
\label{sec:appendix_figures}

\begin{figure*}[t]
  \centering
  \includegraphics[width=\linewidth]{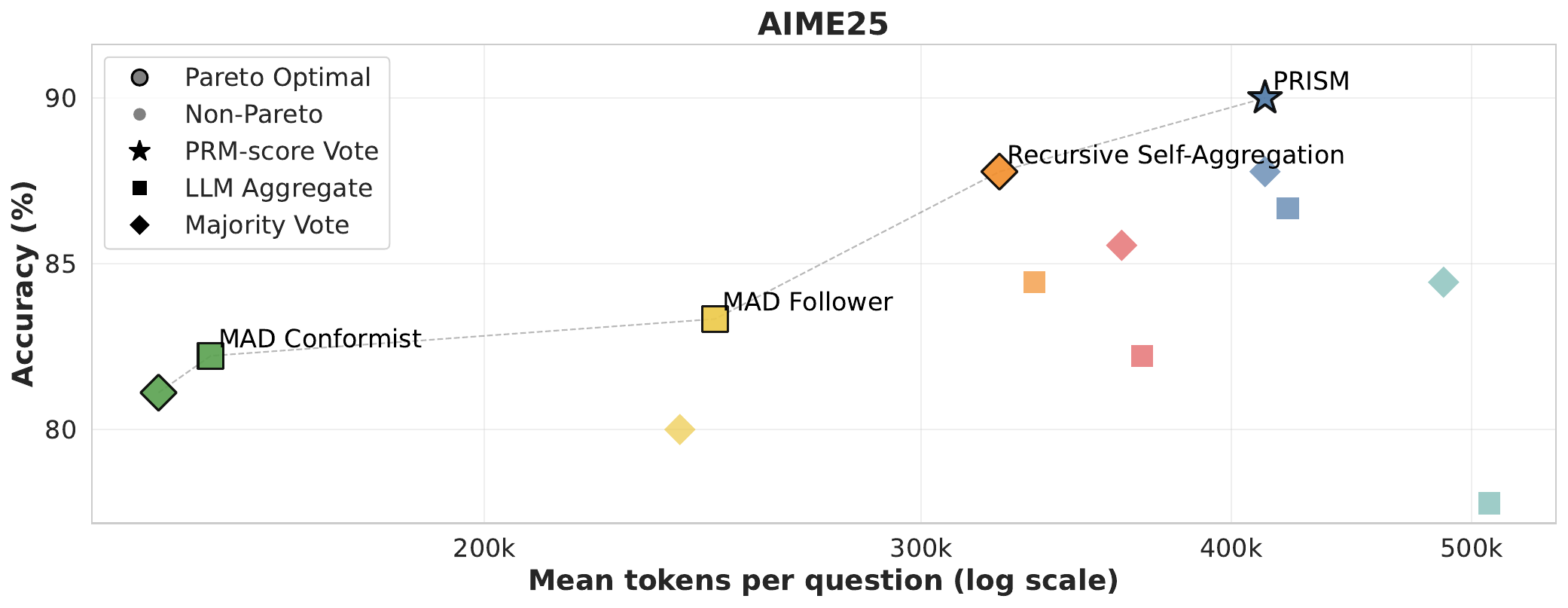}
  \caption{\textbf{Compute--accuracy tradeoff on AIME25.} Each point represents a method configuration (enhancement + aggregation). The x-axis shows mean token usage per question, and the y-axis shows final accuracy. \method achieves the highest accuracy among the evaluated methods.}
  \label{fig:pareto_aime25}
\end{figure*}

\begin{figure*}[t]
  \centering
  \includegraphics[width=\linewidth]{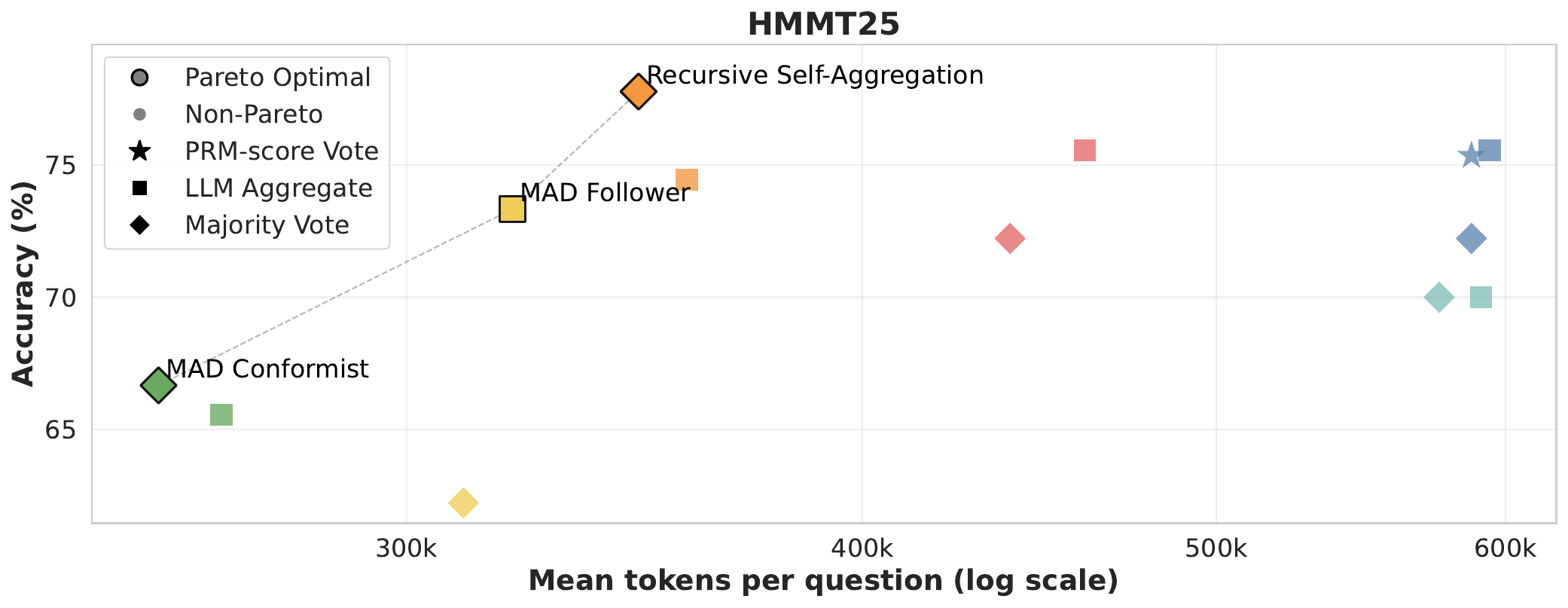}
    \caption{\textbf{Compute--accuracy tradeoff on HMMT25.} Each point represents a method configuration (enhancement + aggregation). The x-axis shows mean token usage per question, and the y-axis shows final accuracy.}
  \label{fig:pareto_hmmt25}
\end{figure*}

\begin{figure*}[t]
  \centering
  \includegraphics[width=0.8\linewidth]{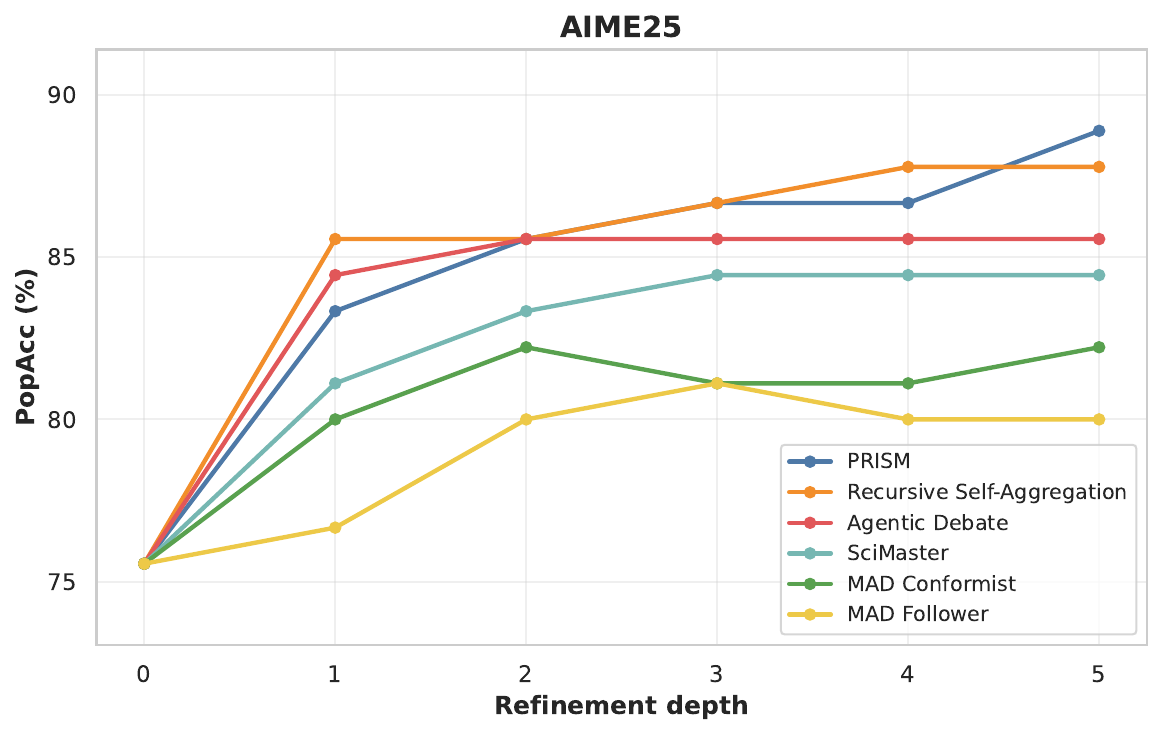}
    \caption{\textbf{Population quality vs.\ refinement depth on AIME25.} Non-PRM-based methods often oscillate or degrade with depth, whereas \method{} exhibits stable upward population dynamics.}
  \label{fig:popacc_depth_aime25}
\end{figure*}

\begin{figure*}[t]
  \centering
  \includegraphics[width=0.8\linewidth]{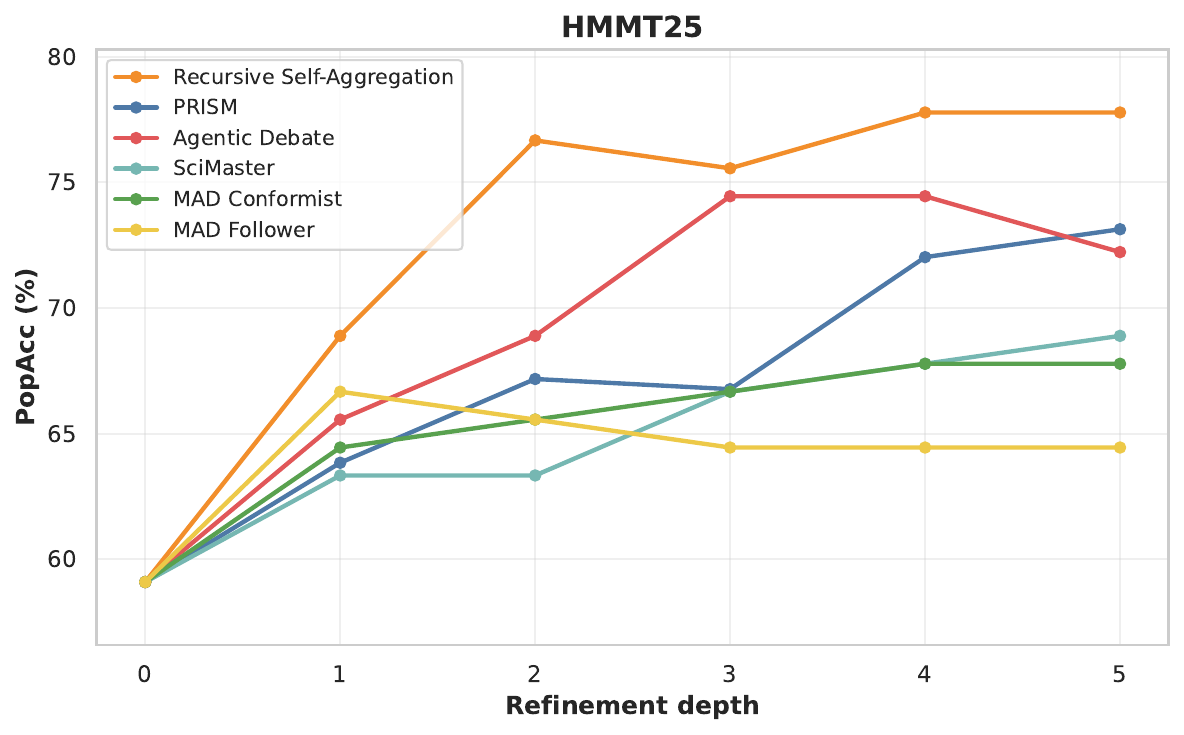}
  \caption{\textbf{Population quality vs.\ refinement depth on HMMT25.} Non-PRM-based methods often oscillate or degrade with depth, whereas \method{} exhibits stable upward population dynamics.}
  \label{fig:popacc_depth_hmmt25}
\end{figure*}

\begin{figure*}[t]
  \centering
  \includegraphics[width=\linewidth]
  {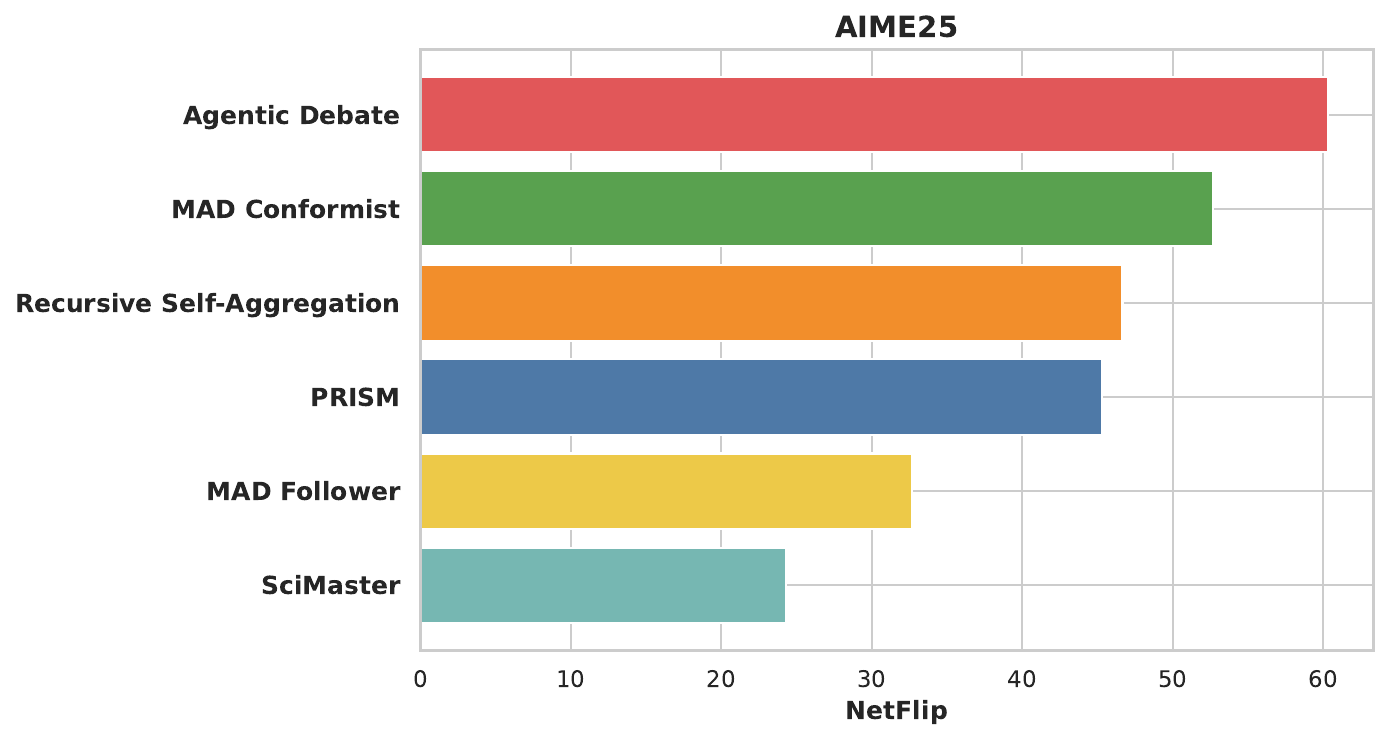}
  \caption{\textbf{Directional correction across enhancement depth on AIME25.} Bars show NetFlip aggregated over depth and averaged across questions. Positive values indicate more incorrect$\rightarrow$correct than correct$\rightarrow$incorrect transitions, reflecting genuine correction rather than ``random-walk'' reshuffling.}
  \label{fig:netflip_aime25}
\end{figure*}

\begin{figure*}[t]
  \centering
  \includegraphics[width=\linewidth]{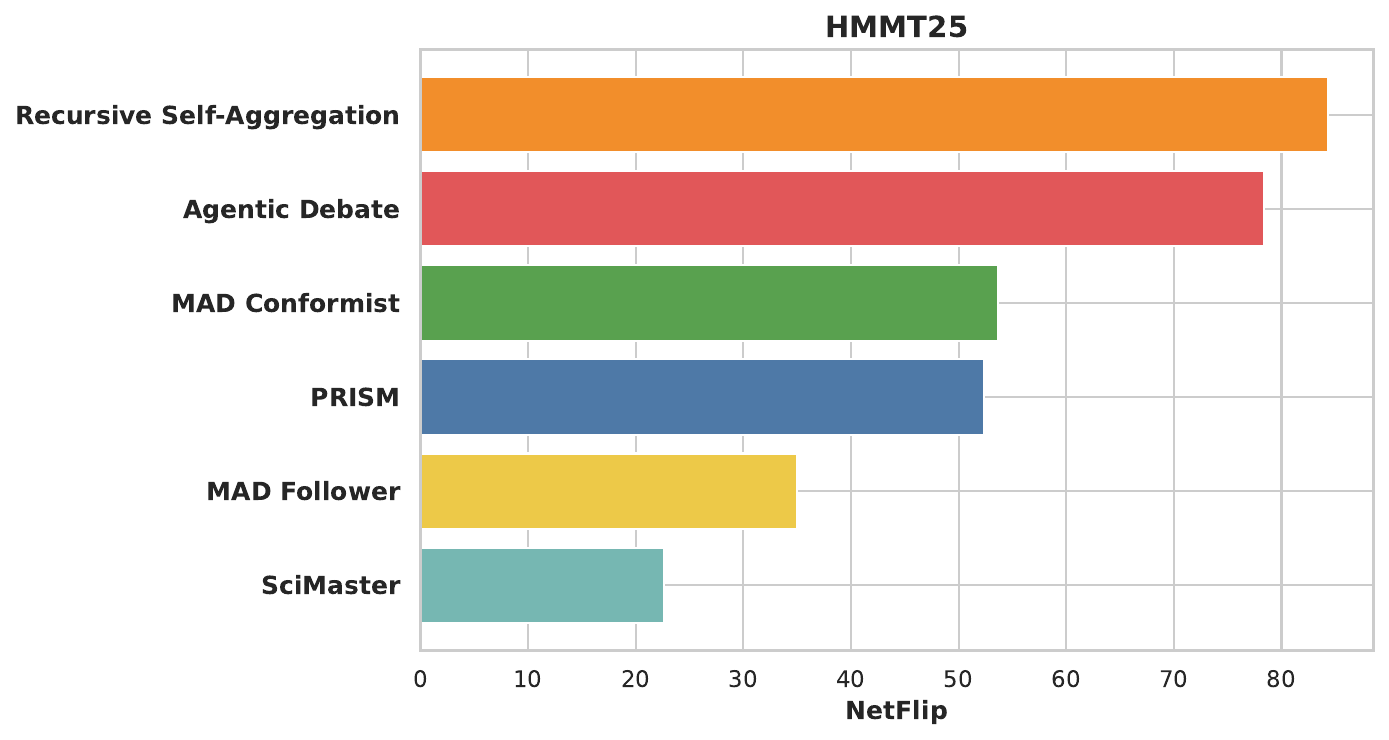}
  \caption{\textbf{Directional correction across enhancement depth on HMMT25.} Bars show NetFlip aggregated over depth and averaged across questions. Positive values indicate more incorrect$\rightarrow$correct than correct$\rightarrow$incorrect transitions, reflecting genuine correction rather than ``random-walk'' reshuffling.}
  \label{fig:netflip_hmmt25}
\end{figure*}

\begin{figure*}[t]
  \centering
  \includegraphics[width=\linewidth]{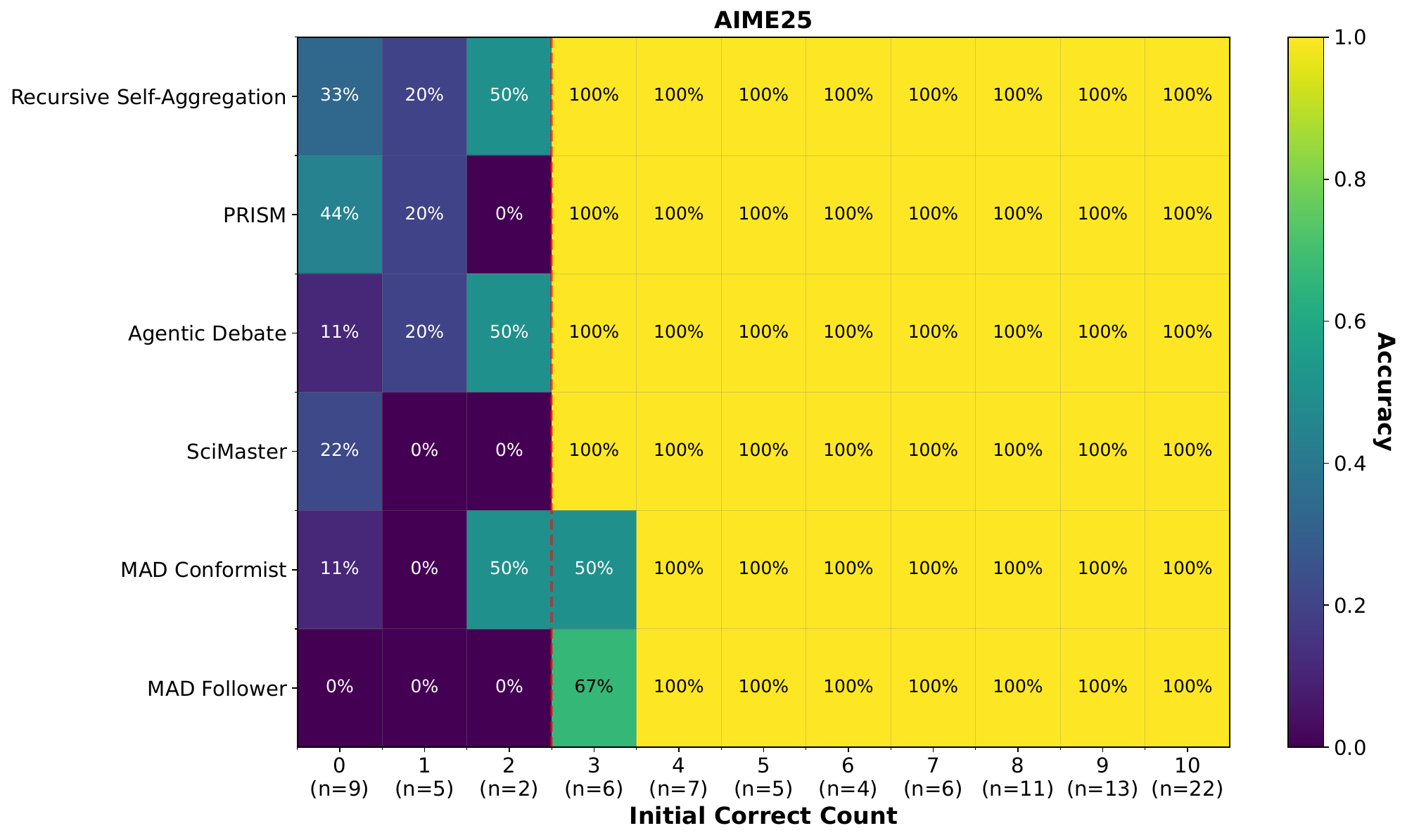}
  \caption{\textbf{AIME25 performance conditioned on initial population quality.} Each column groups problems by how many correct solutions appear in the initial candidate pool, and each cell reports the final accuracy after refinement and majority voting. While most methods struggle when few correct candidates are present at the start, \method{} maintains substantially higher accuracy in these low-correctness regimes, which shows stronger resilience to weak initial populations.}
  \label{fig:heatmap_aime_pop}
\end{figure*}

\begin{figure*}[t]
  \centering
  \includegraphics[width=\linewidth]{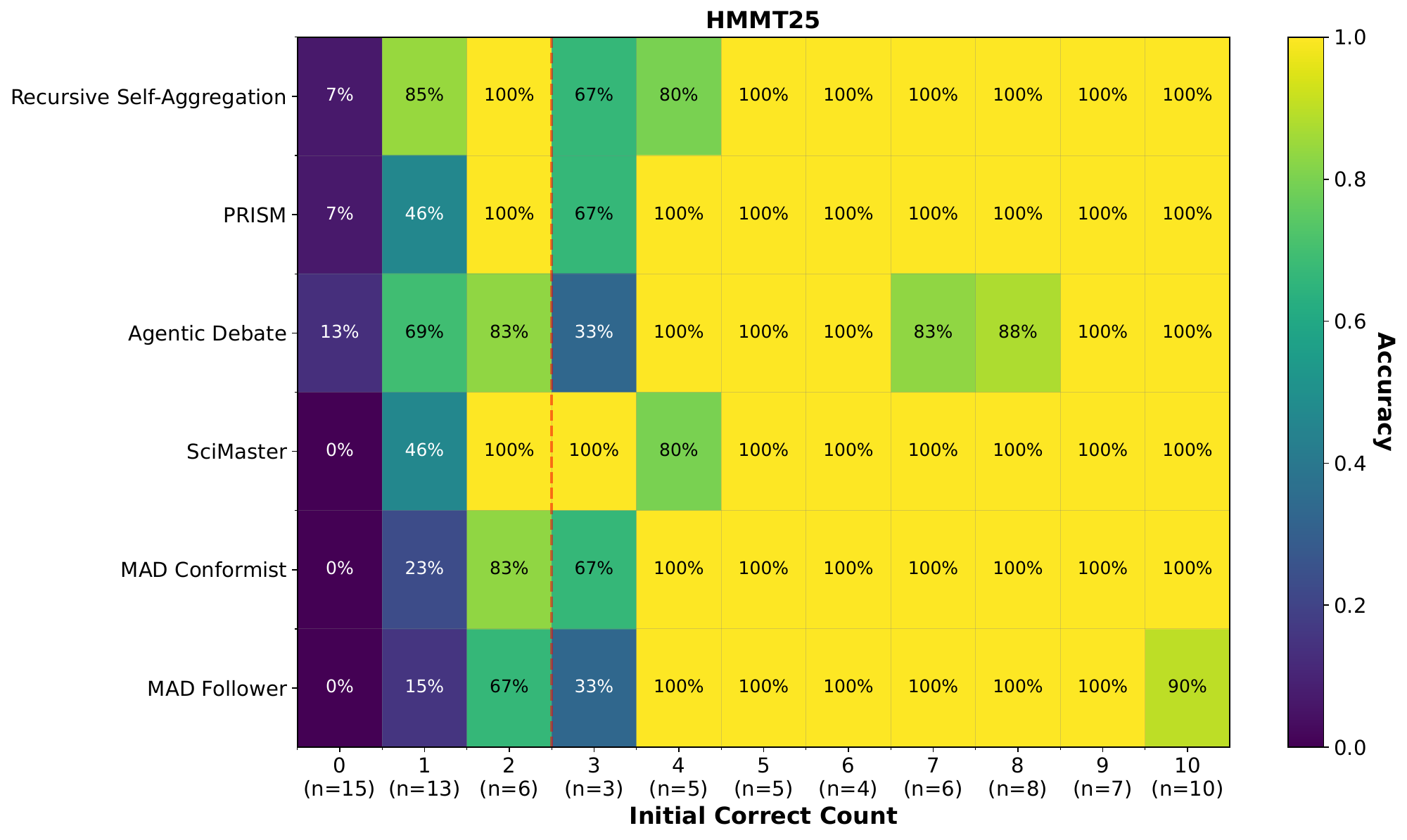}
  \caption{\textbf{HMMT25 performance conditioned on initial population quality.} Each column groups problems by how many correct solutions appear in the initial candidate pool, and each cell reports the final accuracy after refinement and majority voting. While most methods struggle when few correct candidates are present at the start, \method{} maintains substantially higher accuracy in these low-correctness regimes, which shows stronger resilience to weak initial populations.}
  \label{fig:heatmap_hmmt_pop}
\end{figure*}

\begin{figure*}[t]
  \centering
  \includegraphics[width=\linewidth]{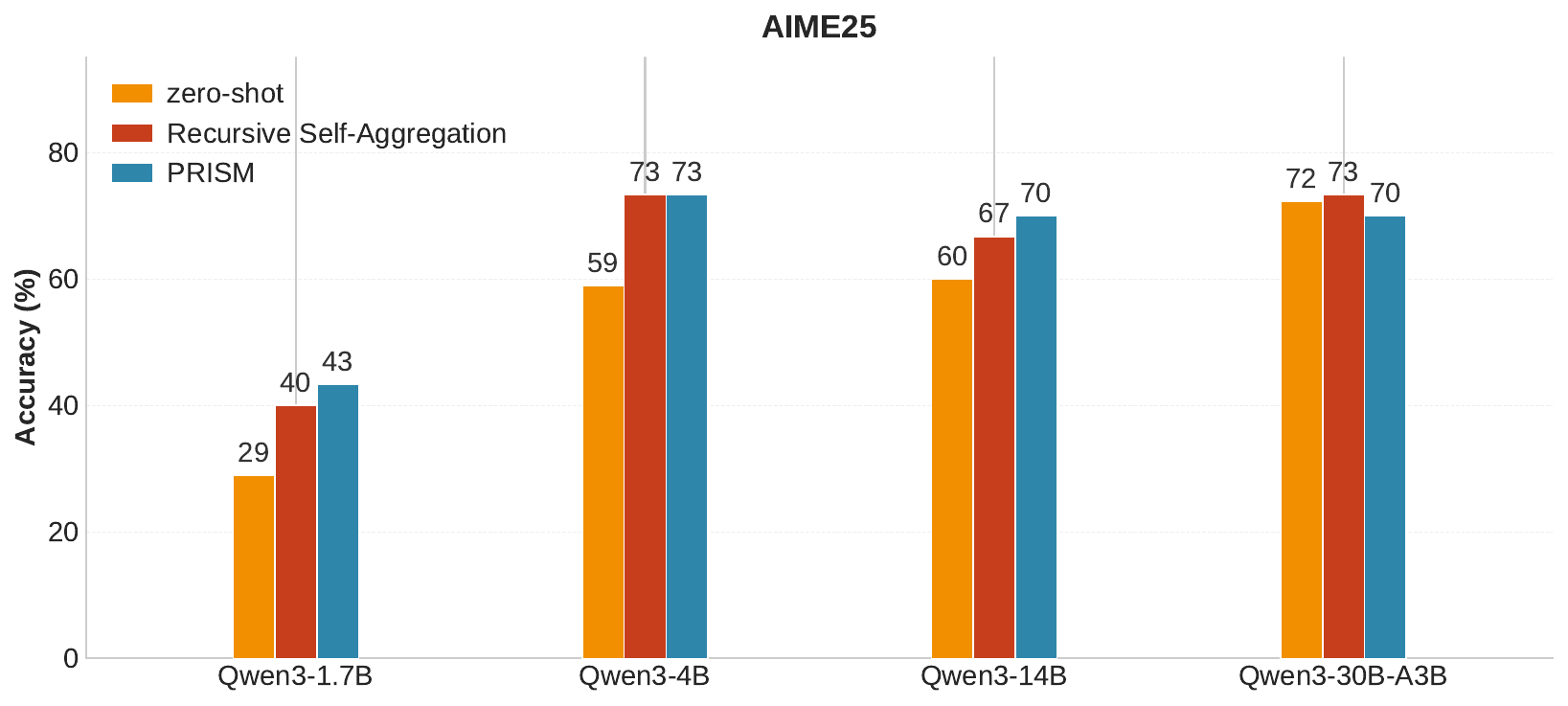}
  \caption{\textbf{AIME25 results on Qwen3 across model sizes.} The bar chart compares zero-shot, Recursive Self-Aggregation, and \method across Qwen3-1.7B, 4B, 14B, and 30B-A3B. \method{} consistently improves over zero-shot performance across model sizes, with the largest gains observed for smaller models.}
  \label{fig:qwen_model_size_aime}
\end{figure*}

\begin{figure*}[t]
  \centering
  \includegraphics[width=\linewidth]{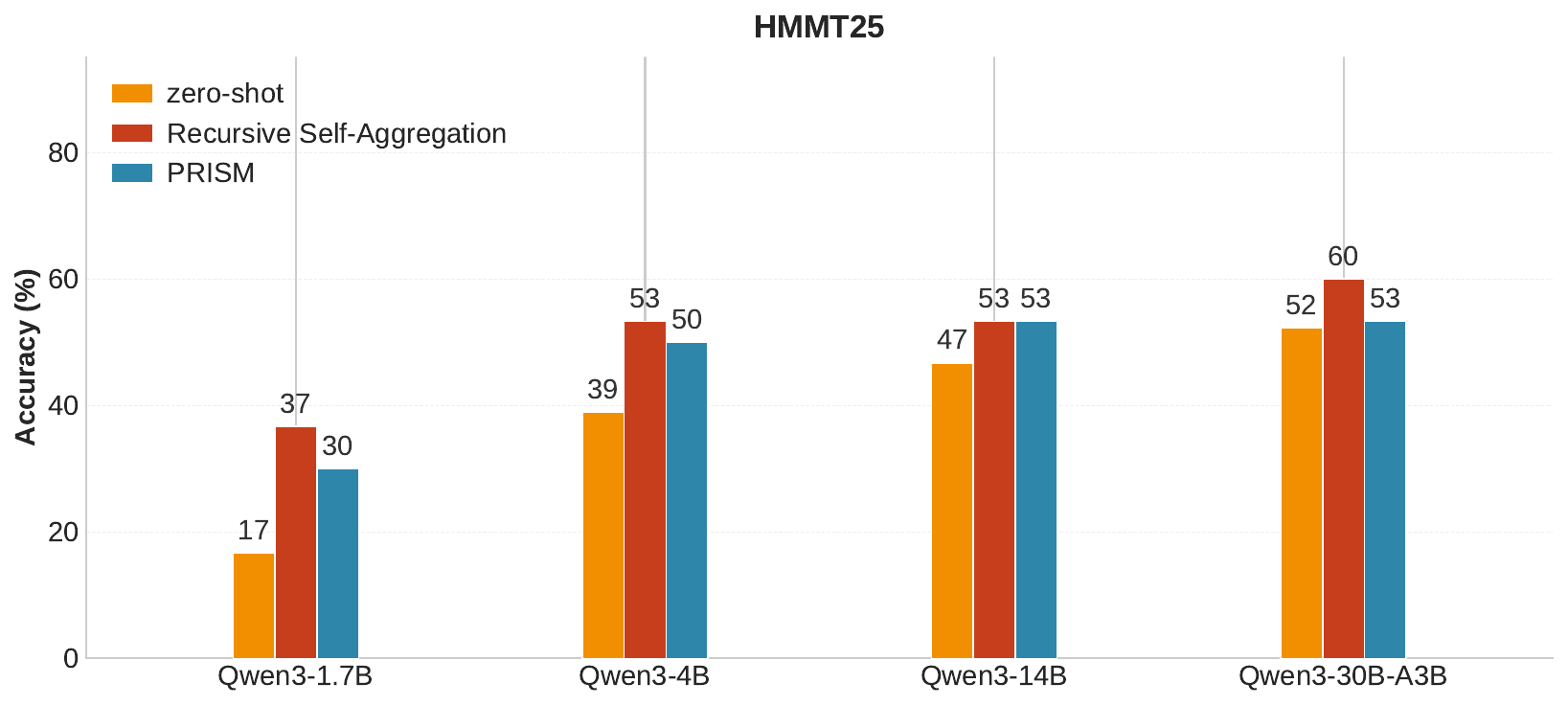}
  \caption{\textbf{HMMT25 results on Qwen3 across model sizes.} The bar chart compares zero-shot, Recursive Self-Aggregation, and \method across Qwen3-1.7B, 4B, 14B, and 30B-A3B. \method{} consistently improves over zero-shot performance across model sizes, with the largest gains observed for smaller models.}
  \label{fig:qwen_model_size_hmmt}
\end{figure*}

\begin{figure*}[t]
  \centering
  \includegraphics[width=\linewidth]{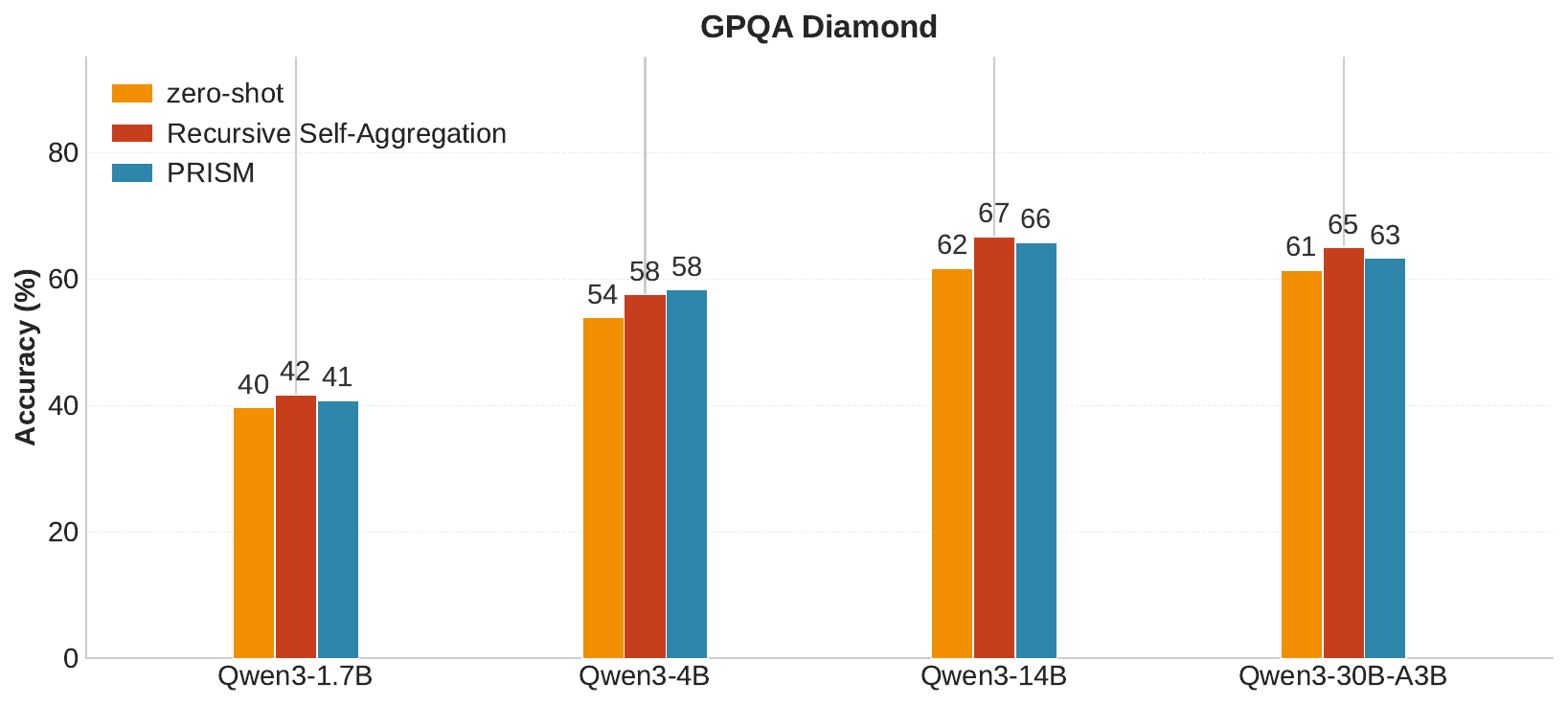}
  \caption{\textbf{GPQA Diamond results on Qwen3 across model sizes.} The bar chart compares zero-shot, Recursive Self-Aggregation, and \method across Qwen3-1.7B, 4B, 14B, and 30B-A3B. \method{} consistently improves over zero-shot performance across model sizes, with the largest gains observed for smaller models.}
  \label{fig:qwen_model_size_gpqa}
\end{figure*}

\begin{figure*}[t]
  \centering
  \includegraphics[width=\linewidth]{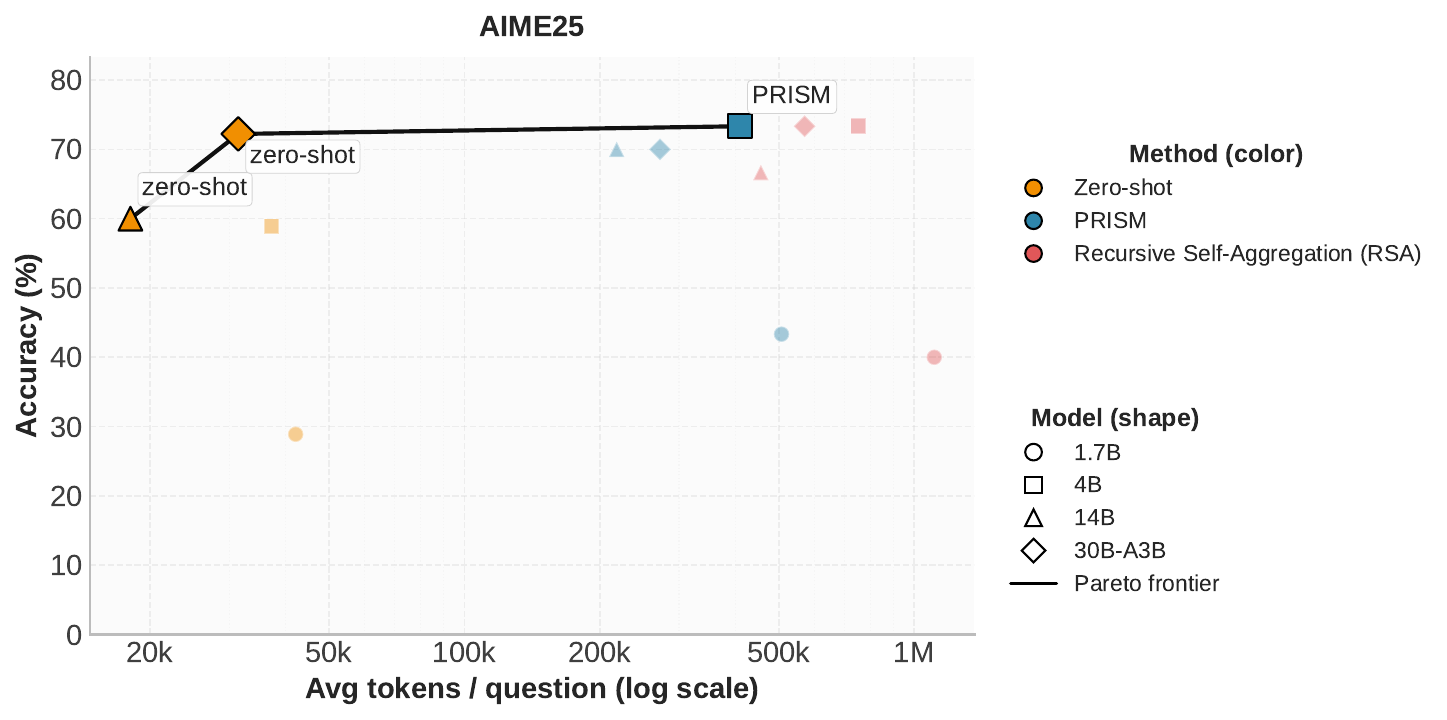}
  \caption{\textbf{AIME25 compute–accuracy tradeoff on Qwen3.} Each point corresponds to a method (color) and model size (marker shape), with the x-axis showing average tokens per question (log scale) and the y-axis showing accuracy.}
  \label{fig:qwen_pareto_aime}
\end{figure*}

\begin{figure*}[t]
  \centering
  \includegraphics[width=\linewidth]{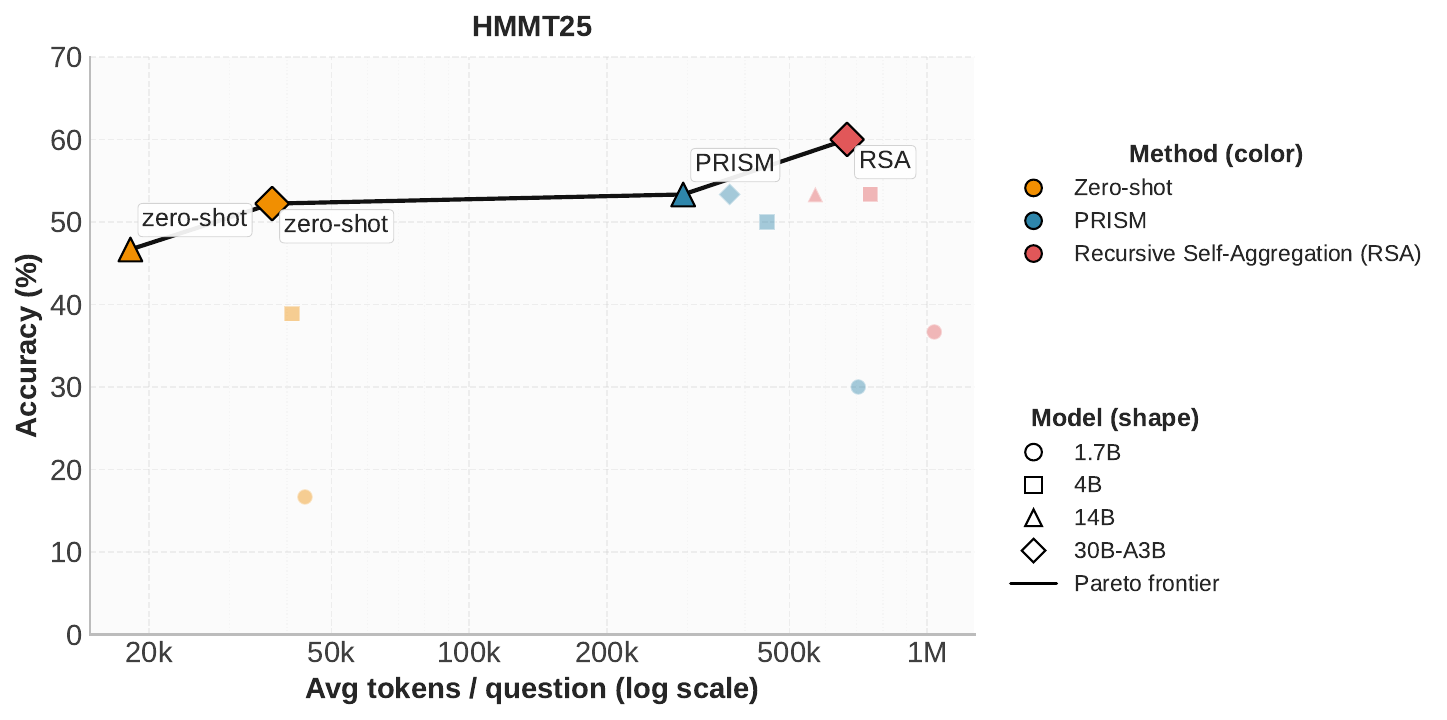}
  \caption{\textbf{HMMT25 compute–accuracy tradeoff on Qwen3.} Each point corresponds to a method (color) and model size (marker shape), with the x-axis showing average tokens per question (log scale) and the y-axis showing accuracy.}
  \label{fig:qwen_pareto_hmmt}
\end{figure*}

\begin{figure*}[t]
  \centering
  \includegraphics[width=\linewidth]{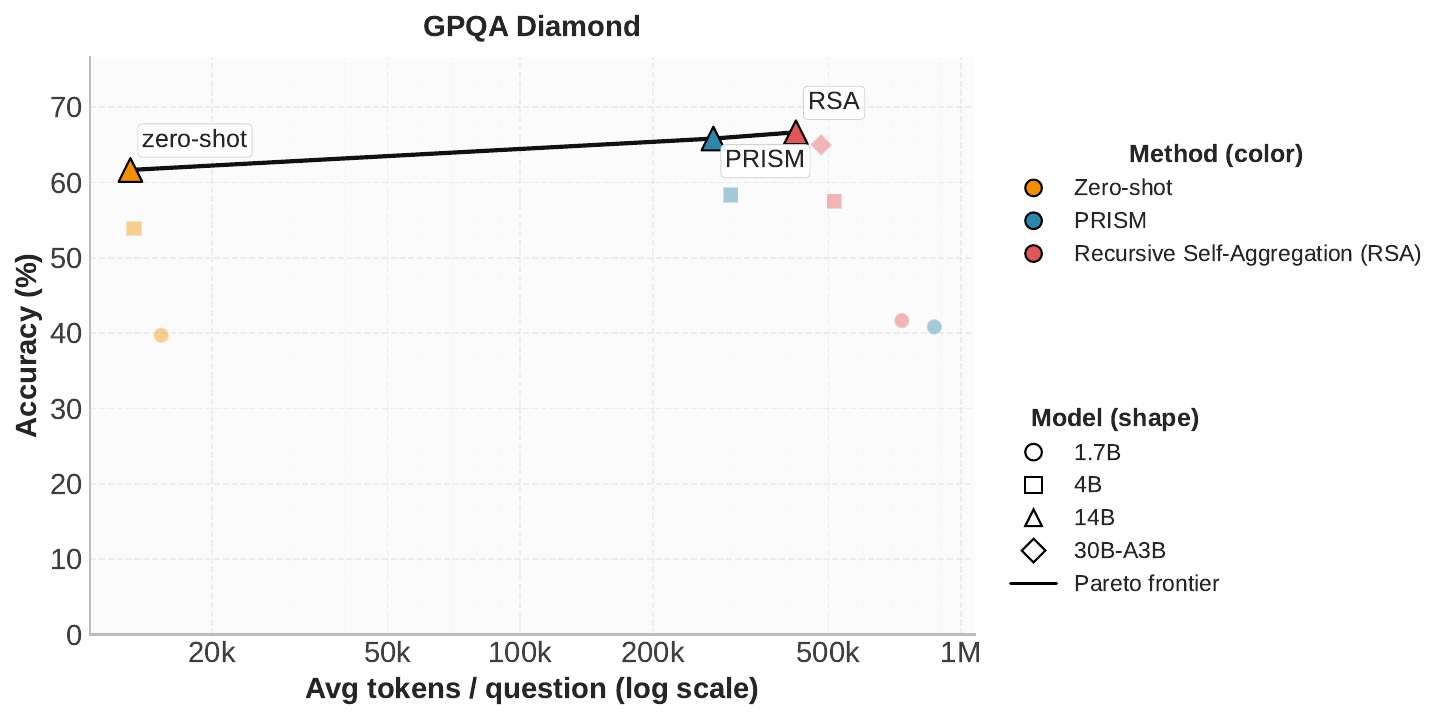}
  \caption{\textbf{GPQA Diamond compute–accuracy tradeoff on Qwen3.} Each point corresponds to a method (color) and model size (marker shape), with the x-axis showing average tokens per question (log scale) and the y-axis showing accuracy.}
  \label{fig:qwen_pareto_gpqa}
\end{figure*}

\begin{figure*}[t]
  \centering
  \includegraphics[width=\linewidth]{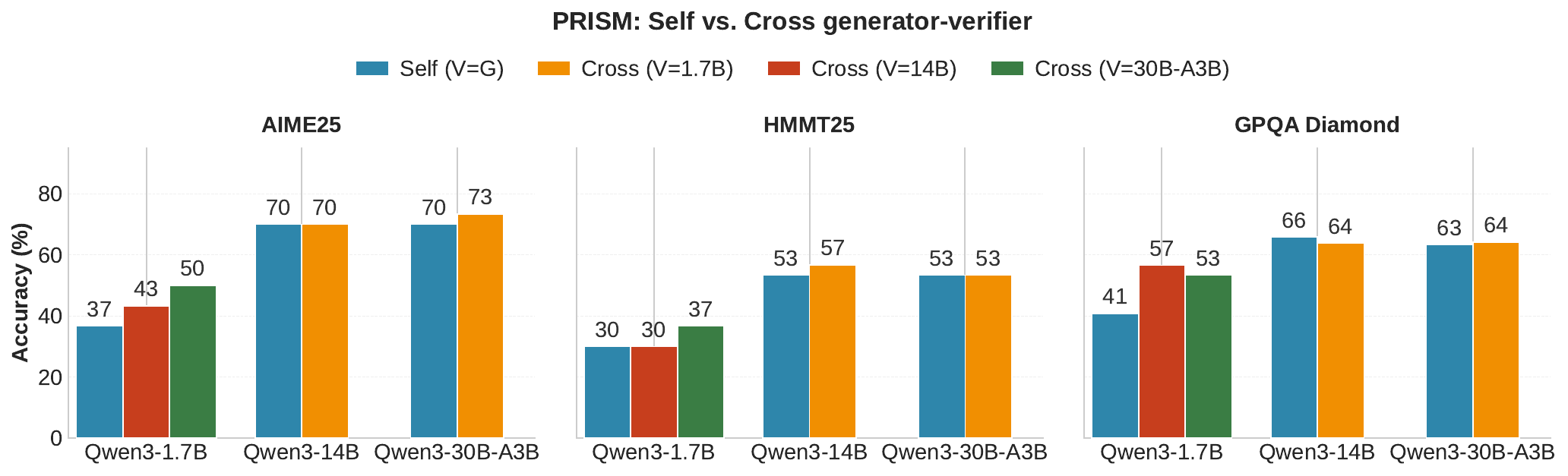}
  \caption{\textbf{Cross generator–verifier scaling on Qwen3.} Bars report \method accuracy when the generator is fixed and the verifier varies, comparing self-verification (V=G) with cross-verification using different verifier sizes. Performance improves as verifier strength increases, with the largest gains occurring when the verifier is larger than the generator.}
  \label{fig:qwen_cross_gen_ver}
\end{figure*}

\begin{figure*}[t]
  \centering
  \includegraphics[width=\linewidth]{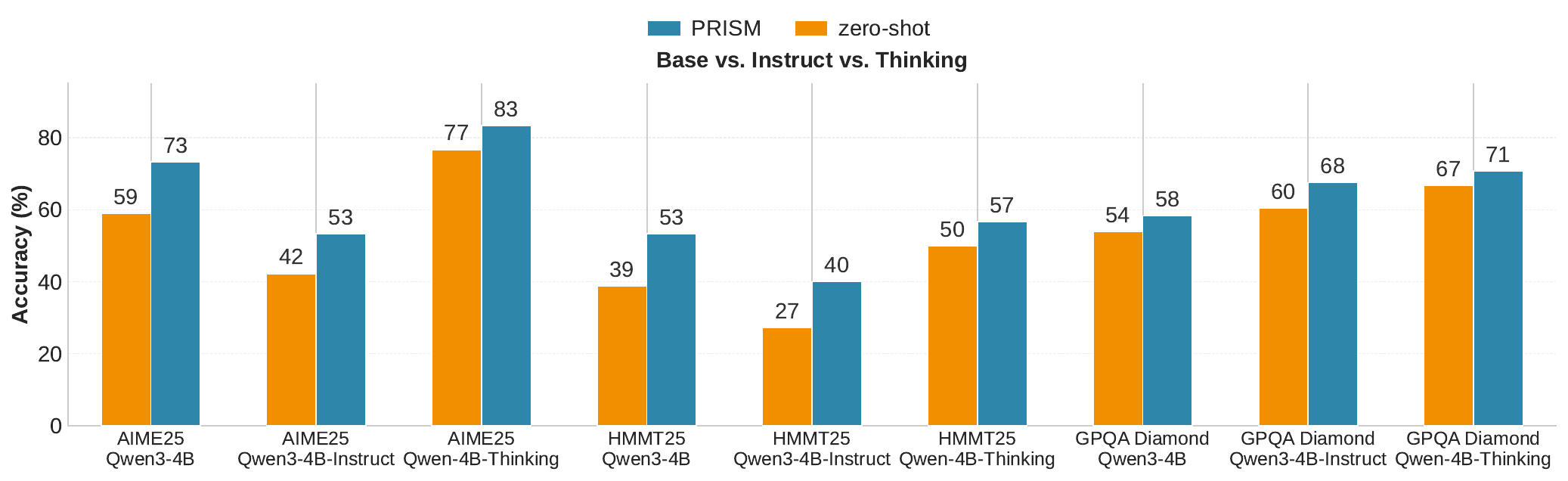}
  \caption{\textbf{Model-type comparison on Qwen3-4B: Base vs. Instruct vs. Thinking variants.} Bars compare zero-shot and \method across closely related variants for each dataset. \method consistently improves performance, with larger gains on weaker variants, thereby narrowing the gap to stronger tuned models.}
  \label{fig:qwen_instruct_think}
\end{figure*}

\end{document}